\documentclass[journal]{IEEEtran}

\usepackage{algorithm}
\usepackage{multirow}
\usepackage{booktabs}
\usepackage{mathrsfs}
\usepackage{cite}
\usepackage{amsmath,amssymb,amsfonts}
\usepackage{algorithmic}
\usepackage{graphicx}
\usepackage{textcomp}
\usepackage{tikz}
\usepackage{adjustbox}
\usepackage{newfloat}
\usepackage{CJKutf8}
\usepackage{listings}
\usepackage{orcidlink}
\usepackage{xcolor}
\definecolor{themeblue}{RGB}{57, 162, 219}
\definecolor{themegreen}{RGB}{87, 204, 153}
\definecolor{forestgreen}{RGB}{47, 159, 87}
\definecolor{link}{RGB}{75, 166, 154}

\usepackage{url}

\newcommand{\cmark}{\color{forestgreen}\ding{51}}%
\newcommand{\xmark}{\color{red}\ding{55}}%
\usepackage{pifont}
\newcommand{\sub}[1]{\textcolor{red}{#1}}
\newcommand{\del}[1]{\textcolor{blue}{#1}}
\newcommand{\ins}[1]{\textcolor{green}{#1}}
\hypersetup{hidelinks=true}

\ifCLASSINFOpdf
\else
\fi
\begin{document}

%
\title{DESign: Dynamic Context-Aware Convolution  and Efficient Subnet Regularization for Continuous Sign Language Recognition}
%
%
%

\author{Sheng Liu~\orcidlink{0000-0001-8082-0903}, \IEEEmembership{Member,~IEEE}, Yiheng Yu, Yuan Feng$^{\dagger}$~\orcidlink{0000-0003-2563-8724}, \IEEEmembership{Member,~IEEE}, Min Xu, Zhelun Jin, Yining Jiang, Tiantian Yuan~\orcidlink{0000-0002-2532-3514}, \IEEEmembership{Member,~IEEE}
\thanks{Sheng Liu, Yiheng Yu, Min Xu, Zhelun Jin, Yining Jiang are with the College of Computer Science, Zhejiang University of Technology, Hangzhou 310014, China.\\
Yuan Feng is with the School of Mathematical Sciences, Zhejiang University of Technology, Hangzhou 310014, China (e-mail: fy@ieee.org).\\
Tiantian Yuan is with the Technical College for the Deaf, Tianjin University of Technology, Tianjin 300384, China.\\

$\dagger$ Yuan Feng is the corresponding author.}}

%
%

\markboth{Journal of \LaTeX\ Class Files,~Vol.~14, No.~8, August~2015}%
{Shell \MakeLowercase{\textit{et al.}}: Bare Demo of IEEEtran.cls for IEEE Journals}
%



\maketitle

\begin{abstract}
Current continuous sign language recognition (CSLR) methods struggle with handling diverse samples. Although dynamic convolutions are ideal for this task, they mainly focus on spatial modeling and fail to capture the temporal dynamics and contextual dependencies. To address this, we propose DESign, a novel framework that incorporates Dynamic Context-Aware Convolution (DCAC) and Subnet Regularization Connectionist Temporal Classification (SR-CTC). DCAC dynamically captures the inter-frame motion cues that constitute signs and uniquely adapts convolutional weights in a fine-grained manner based on contextual information, enabling the model to better generalize across diverse signing behaviors and boost recognition accuracy. Furthermore, we observe that existing methods still rely on only a limited number of frames for parameter updates during training, indicating that CTC learning overfits to a dominant path. To address this, SR-CTC regularizes training by applying supervision to subnetworks, encouraging the model to explore diverse CTC alignment paths and effectively preventing overfitting. A classifier-sharing strategy in SR-CTC further strengthens multi-scale consistency. Notably, SR-CTC introduces no inference overhead and can be seamlessly integrated into existing CSLR models to boost performance. Extensive ablations and visualizations further validate the effectiveness of the proposed methods. Results on mainstream CSLR datasets (i.e., PHOENIX14, PHOENIX14-T, CSL-Daily) demonstrate that DESign achieves state-of-the-art performance.

\end{abstract}


%
\IEEEpeerreviewmaketitle

\section{Introduction}
\label{introduction}
Sign language is a visual language widely used in daily communication among the deaf community. Unlike spoken languages, it conveys meaning through both manual components (hand and arm gestures) and non-manual components (facial expressions, head movements, and body postures) \cite{dreuw2007speech,ong2005automatic}, following a unique grammar and syntax \cite{Reagan_2007,alyami2024reviewing} which make it challenging to master in a short time. To facilitate communication, isolated sign language recognition (ISLR) \cite{freeman1995orientation,sun2013discriminative,tunga2021pose,hu2021signbert} simplifies the problem by identifying individual sign gestures in short video clips and mapping them to corresponding glosses \footnote{Gloss is the lexical unit of annotated continuous sign language.}. In contrast, continuous sign language recognition (CSLR) \cite{vac,smkd,corrnet,twostream,tcnet} aims to recognize gloss sequences from videos containing successive signs, making it a more complex yet valuable research area. In this paper, we focus on CSLR and its advancements.

Traditional CSLR methods lack the ability to dynamically adapt their weights to input variations \cite{corrnet,corrnet+}, which can compromise their robustness when faced with diverse or unseen inputs. To overcome this limitation, enhancing the model’s adaptability to varying input patterns is essential. Dynamic convolution emerges as a promising solution, as it enables input-dependent feature adaptation \cite{jia2016dynamic,chen2020dynamic}. However, applying this mechanism to CSLR introduces several challenges. First, existing dynamic convolution methods in the visual domain \cite{odconv, tada1, tada2} are primarily designed for spatial modeling. This frame-isolated processing fails to capture cross-frame signals such as motion trajectories, which are proven essential for accurate sign recognition in CSLR \cite{corrnet, corrnet+}. Moreover, sign language exhibits strong temporal contextual dependencies \cite{cvsign, olmd} that are distinct from those in general video tasks. However, current methods can only coarsely adjust their weights based on temporal features \cite{tam, tada1, tada2}, making it difficult to achieve strong context-awareness and limiting their performance when transferred to CSLR. As a result, the inability to fully exploit inter-frame dynamics and fine-grained temporal context significantly limits the effectiveness of existing dynamic convolution methods when applied to CSLR.

\begin{figure*}
\centering
  \includegraphics[width=0.75\textwidth]{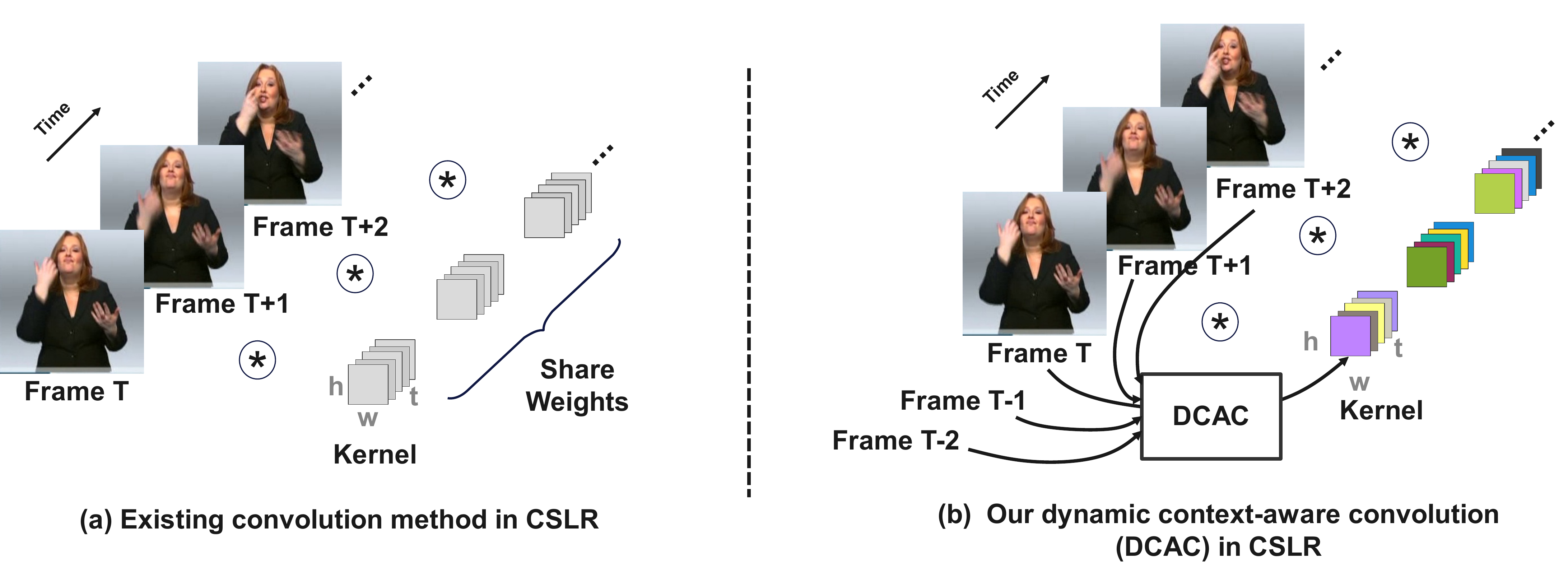}
  \caption{\textbf{A comparison between (a) existing convolution methods in CSLR and (b) our Dynamic Context-Aware Convolution (DCAC).} Existing convolutional approaches in CSLR \cite{sen, corrnet, corrnet+} share the same kernel across frames and lack the ability to dynamically adjust weights based on input samples. In contrast, DCAC adaptively adjusts the convolution kernel for each frame according to its contextual information, exhibiting strong capabilities in both sample adaptability and context awareness.}
  
  \label{fig:img1}
\end{figure*}


Connectionist Temporal Classification (CTC) remains the dominant alignment strategy in contemporary CSLR models. While it facilitates end-to-end training, CTC also introduces several notable limitations. Prior studies \cite{dnf,Pu2019IterativeAN} have shown that it weakens the discriminative power of feature extractors and suffers from the spike phenomenon, wherein effective supervision is concentrated on a limited subset of frames. To mitigate these issues, VAC \cite{vac} introduces auxiliary CTC losses and self-distillation to enhance feature learning, while SMKD \cite{smkd} alleviates spikes by leveraging gloss-level segmentation. RadialCTC \cite{radialctc} further addresses this challenge by constraining the alignment path to suppress extreme peaking. Despite these advancements, the problem still remains largely unresolved. As illustrated in Fig. \ref{fig:img4}, baseline methods—including \cite{vac,smkd,tlp}—continue to rely on a limited number of frames for gradient updates. This results in a persistent risk of overfitting to the dominant alignment path and leads to near-stagnation in model optimization. Therefore, developing more effective alignment mechanisms remains a critical research direction in CSLR.

To address the aforementioned challenges, we propose DESign. Specifically, we first design a \textbf{Dynamic Context-Aware Convolution (DCAC)} for CSLR. Unlike existing dynamic convolutions, DCAC adapts weights frame-by-frame while also modeling cross-frame dependencies to capture motion transitions inherent in sign language. By explicitly extracting the temporal context as a guiding signal for kernel generation, DCAC produces highly context-aware convolutional weights. This significantly improves the model’s ability to capture subtle temporal variations and semantic continuity, leading to more accurate recognition of continuous and ambiguous signs. To highlight the distinct advantages of DCAC in CSLR, we compare the core operating mechanisms of existing convolutions in CSLR and DCAC in Fig. \ref{fig:img1}. Furthermore, inspired by Intermediate CTC \cite{lee2021intermediate}, we introduce \textbf{Subnet Regularization CTC (SR-CTC)} to alleviate overfitting. SR-CTC applies auxiliary losses to intermediate subnetworks during training, effectively regularizing the learning process and encouraging the model to explore more optimal alignment paths. Additionally, we incorporate a \textbf{classifier-sharing strategy} within SR-CTC to promote implicit semantic alignment across multi-scale features. Notably, SR-CTC significantly improves performance while being simple to implement and introducing no additional inference overhead. \textbf{In a nutshell},  our contributions are summarised as threefold:
\begin{itemize}
    \item[$\bullet$] A novel Dynamic Context-Aware Convolution (DCAC) is proposed for CSLR. It integrates per-frame contextual adaptation with cross-frame dependency modeling, effectively aligning dynamic convolution with the specific demands of CSLR and yielding superior performance.
    
    \item[$\bullet$] We propose Subnet Regularization CTC (SR-CTC) to address the CTC spike issue by regularizing subnetwork training, effectively preventing overfitting. A classifier-sharing strategy further promotes implicit alignment of multi-scale sign features. SR-CTC is lightweight and can be seamlessly integrated into existing CSLR models without introducing any inference overhead.
   
    \item[$\bullet$] Extensive ablations and visualizations demonstrate the effectiveness of the proposed methods. Validation results on the PHOENIX14, PHOENIX14-T, and CSL-Daily show that DESign achieves state-of-the-art performance without using any additional cues or auxiliary datasets.
\end{itemize}

\vspace{-4pt}

\section{Related Work}
\subsection{Continuous Sign Language Recognition}

Continuous sign language recognition (CSLR) \cite{cosign,cvtslr,vac,stmc,sen,corrnet} aims to learn the mapping from video sequences to word-level annotations under weakly supervised conditions, where only sentence-level annotations are provided. CSLR faces two major challenges: feature extraction and sequence alignment. For feature extraction, early methods rely on handcrafted features such as HOG \cite{hog} and Fourier descriptors \cite{fourier}, but these have been completely replaced by deep learning methods. Recent methods \cite{vac,smkd,corrnet,sen,c2slr} typically decouple spatial and temporal modeling—using 2D CNNs \cite{vac,smkd,corrnet,sen} or Transformers \cite{c2st} for frame-wise feature extraction, followed by 1D CNNs and BiLSTMs for local and global temporal modeling. Sequence alignment in CSLR is similar to automatic speech recognition \cite{jointctc,autoasr,ma2021end,nozaki2021relaxing}, but due to limited data availability, only HMM- \cite{koller2017re,koller2019weakly} or CTC-based \cite{vac,corrnet,twostream} methods are feasible, with CTC being the most widely used due to its simplicity and efficiency. Overall, decoupled spatiotemporal modeling (2D CNN+1D CNN+BiLSTM) and CTC are the mainstream methods in current CSLR.

Recently, numerous methods have been proposed to improve CSLR baseline architectures and address existing challenges. To alleviate the impact of CTC spikes, VAC \cite{vac} introduces two regularization losses, while SMKD \cite{smkd} further enhances visual-text feature discrimination through a classifier-sharing strategy. In addition, recognizing the limitations of frame-wise feature extraction, SEN \cite{sen}, CorrNet \cite{corrnet}, and TCNet \cite{tcnet} design various cross-frame attention mechanisms to better capture motion trajectories. Moreover, several approaches explore modalities beyond RGB to improve recognition performance. CVT-SLR \cite{cvtslr}, CTCA \cite{ctca}, and C$^2$ST \cite{c2st} incorporate textual information for auxiliary training, while C$^2$SLR \cite{c2slr}, TwoStream \cite{twostream}, and SignVTCL \cite{signvtcl} introduce additional visual cues (e.g., keypoints, optical flow) to enhance visual representation. However, these multi-cue methods often entail higher training costs, more complex architectures, and increased inference latency, significantly limiting their practicality in real-world CSLR applications. In this paper, we aim to enhance CSLR by integrating improved dynamic convolution to strengthen the model’s capacity, along with optimized supervision to mitigate overfitting. As a result, the proposed DESign achieves new state-of-the-art performance using only RGB input.

\begin{figure*}
 \centering
  \includegraphics[width=0.8\textwidth]{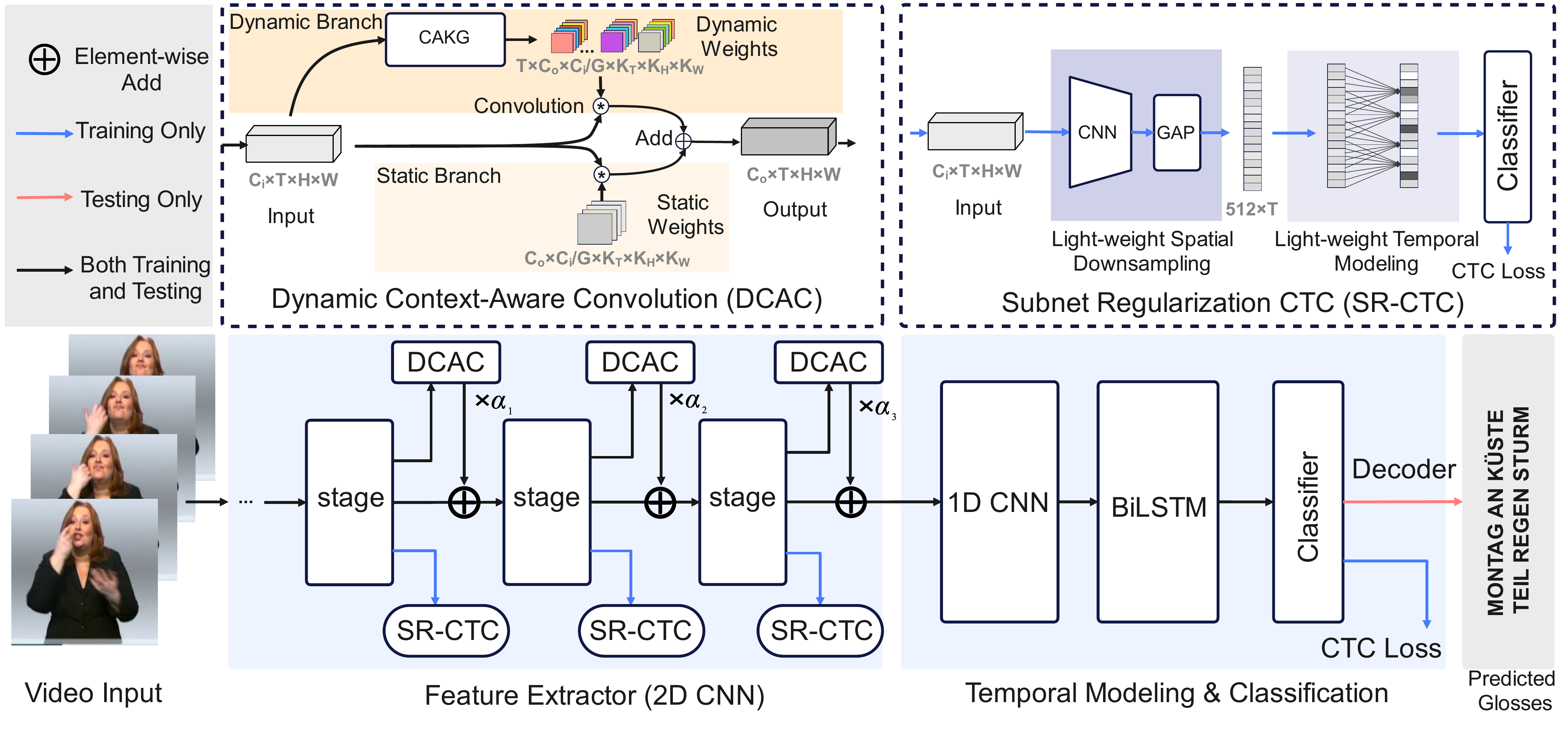}
  \caption{An overview of DESign. The architecture comprises a frame-level feature extractor (2D CNN), followed by a 1D CNN and a two-layer BiLSTM for capturing local and global temporal dependencies. A classifier generates the output representations, which are decoded into gloss sequences during inference. Dynamic Context-Aware Convolution (DCAC) is inserted after each stage to enhance adaptability and contextual perception. SR-CTC treats the network segments preceding each stage as subnetworks and applies auxiliary CTC losses during training to regularize optimization, effectively mitigating gradient vanishing and overfitting.
}  
  
  \label{fig:img2}
\end{figure*}
\begin{table}[t!]
\centering
\caption{Comparison with Existing Dynamic Convolutions. Frame-wise adaptive indicates the ability to generate and adjust weights on a per-frame basis, while Temporal adaptive refers to the ability to adjust weights according to temporal variations}
\resizebox{0.49\textwidth}{!}{%
\begin{tabular}{lcccc}
\toprule
\multirow{2}{*}{\textbf{Operations}} 
  & \textbf{Frame-wise} & \textbf{Temporal} & \textbf{Cross-Frame} & \textbf{Context} \\
  & \textbf{adaptive} & \textbf{adaptive} & \textbf{modeling} & \textbf{awareness} \\
\midrule
\multicolumn{5}{c}{\textbf{Image-based Methods}} \\
CondConv \cite{yang2019condconv}     & \cmark & \xmark & \xmark & \xmark \\
DyConv \cite{chen2020dynamic}        & \cmark & \xmark & \xmark & \xmark \\
ODConv \cite{odconv}                 & \cmark & \xmark & \xmark & \xmark \\
Bi-volution \cite{bivolution}         & \cmark & \xmark & \xmark & \xmark \\
\midrule
\multicolumn{5}{c}{\textbf{Video-based Methods}} \\
TAM \cite{tam}                       & \xmark & \cmark & \cmark & \xmark \\
TAda V1 \cite{tada1}        & \cmark & \cmark & \xmark & \xmark \\
TAda V2 \cite{tada2}        & \cmark & \cmark & \xmark & \xmark \\
\textbf{DCAC (ours)}                          & \cmark & \cmark & \cmark & \cmark \\
\bottomrule
\end{tabular}
}
\label{tab:1}
\end{table}
\vspace{-10pt}
\subsection{Dynamic Convolution}
Dynamic convolution adapts kernel weights based on input data \cite{klein2015dynamic,ha2016hypernetworks,jia2016dynamic}, offering greater expressiveness and adaptability. In the image domain, DyConv \cite{chen2020dynamic} and CondConv \cite{yang2019condconv} use attention mechanisms to dynamically assign weights to multiple expert kernels, improving model complexity and performance. Building on this idea, ODConv \cite{odconv} introduces all-dimensional attention to further enhance representational capacity. Bi-volution \cite{bivolution} combines static and dynamic convolutions in a dual-branch architecture, demonstrating improved robustness to noise. In the video domain, TAM \cite{tam} integrates global and local modeling into 1D dynamic convolution to better capture temporal features. However, it lacks frame-wise adaptability and relies on fixed-length inputs, limiting flexibility. TAda V1, V2 \cite{tada1,tada2} addresses this by calibrating 2D convolutional kernels on a per-frame basis using temporal cues, enabling improved adaptation to temporal variation. Nonetheless, TAda primarily focuses on spatial modeling and lacks temporal context awareness—a critical component for effective CSLR, which depends heavily on capturing sequential dependencies and contextual cues. To better align dynamic convolution with the demands of CSLR, we propose DCAC. As shown in Tab.~\ref{tab:1}, compared to existing dynamic convolution methods, DCAC uniquely achieves both frame-wise adaptability and cross-frame modeling with enhanced context awareness, leading to significantly improved performance on CSLR tasks.

\subsection{Connectionist Temporal Classification}
Connectionist Temporal Classification (CTC) \cite{ctc} is widely used to handle sequence length mismatches between inputs and outputs, facilitating end-to-end learning. However, it inevitably introduces several limitations (e.g., CTC spike) that significantly impact performance. As a result, extensive research has focused on addressing these issues. In automatic speech recognition (ASR), Intermediate CTC \cite{lee2021intermediate} applies auxiliary supervision to intermediate layers to regularize training and improve performance at minimal cost, while self-conditioned CTC \cite{nozaki2021relaxing} further alleviates the conditional independence constraint by leveraging intermediate predictions. InterAug CTC \cite{nakagome2022interaug} enhances model robustness by injecting noise into intermediate predictions. To mitigate the peaking phenomenon, CR-CTC \cite{crctc} introduces a siamese network that regularizes CTC distributions from different augmented views. Additionally, joint CTC-attention \cite{yu2021boundary,fan2021cass,han2023knowledge,dong2020cif,hori2017advances} is widely adopted to address CTC’s limitations in capturing long-range dependencies and contextual information. Although these methods have achieved remarkable success in ASR, directly transferring them to CSLR is nontrivial due to significant differences in network architectures, data scales, and modalities. Consequently, the CSLR community has proposed several domain-specific adaptations.  VAC \cite{vac} introduces self-distillation between semantic and visual features to mitigate the peaking problem. Building upon this, SMKD \cite{smkd} further proposes gloss segmentation and weight sharing strategies to alleviate this issue. RadialCTC \cite{radialctc} constrains sequence representations on a hypersphere to enhance alignment quality. C$^2$ST \cite{c2st} conditions the decoding process on predicted gloss sequences, thereby relaxing the conditional independence assumption. Collectively, these improvements have established strong baselines for CSLR.

However, we observe that existing CSLR methods still suffer from CTC spikes, with only a few frames effectively contributing to parameter updates. Inspired by Intermediate CTC \cite{lee2021intermediate}, we propose an efficient solution called SR-CTC to tackle this issue. SR-CTC explores alignment paths using low-level features from intermediate subnetworks, serving as a regularization strategy that effectively prevents the model from overfitting to dominant paths. Additionally, a classifier-sharing mechanism is introduced to promote semantic alignment across features at different scales. Finally, SR-CTC can be seamlessly integrated into existing frameworks in a lightweight manner, leading to improved performance.

\section{Method}

\subsection{Problem Definition and Overview}
CSLR aims to map a video with $T$ frames, $\boldsymbol{V}=\left\{v_{t}\right\}_{t=1}^{T}\in\mathcal{R}^{T\times H\times W\times3}$, to a gloss sequence $\mathcal{G}=\{g_{i}\}_{i=1}^{N}$, where $N$ is the sequence length and  satisfies $N<T/4$. Existing approaches use a 2D-CNN as frame-wise feature extractor to obtain $\boldsymbol{Y} = \{ y_{t} \}_{t=1}^{T} \in \mathcal{R}^{T \times d_v}$, where $d_v$ is the feature dimension. A 1D-CNN is then applied to capture local temporal features, yielding $\boldsymbol{L} = \{ l_{t} \}_{t=1}^{T'} \in \mathcal{R}^{T' \times d_s}$ where $T’$  represents the sequence length and satisfies $N<T'<T/4$, whith $d_s$ as the feature dimension. To further model long-range dependencies, a two-layer BiLSTM is utilized to extract global temporal features, $\boldsymbol{G} = \{ g_{t} \}_{t=1}^{T'} \in \mathcal{R}^{T' \times d_s}$. Finally, the outputs are classified and then optimized with CTC loss.

The overall architecture of DESign is illustrated in Fig. \ref{fig:img2}. The Dynamic Context-Aware Convolution (DCAC)  is inserted after each stage of the feature extractor. Adopting a parallel architecture with dynamic and static branches to balance noise \cite{bivolution}, DCAC centers on introducing Context-Aware Kernel Generation (CAKG) in the dynamic branch, which generates convolutional kernels for each frame based on contextual information, endowing the model with strong context awareness. Meanwhile, SR-CTC applies additional CTC supervision to the outputs of each stage, effectively regularizing the training process. It also introduces Light-weight Spatial Downsampling and Light-weight Temporal Modeling to perform feature downsampling and temporal modeling before CTC loss computation—both of which are only activated during training. In the following sections, we provide a detailed description of the two core innovations of this work: DCAC and SR-CTC.

\vspace{-0.5 pt}

\begin{algorithm}[t]
\caption{\small{Pseudo code of DCAC in a PyTorch-like style.}}
\label{alg:code}
\definecolor{codeblue}{rgb}{0.25,0.5,0.5}
\lstset{
	backgroundcolor=\color{white},
	basicstyle=\fontsize{7.2pt}{7.2pt}\ttfamily\selectfont,
	columns=fullflexible,
	breaklines=true,
	captionpos=b,
	commentstyle=\fontsize{7.2pt}{7.2pt}\color{codeblue},
	keywordstyle=\fontsize{7.2pt}{7.2pt},
}
\vskip -0.075in
\begin{lstlisting}[language=python]
# B: batch size, T: temporal length, H: height, W: width 
# C: channels, G: number of groups, n: number of experts
# kt, kh, kw: kernel size (temporal, height, width)
# r: reduction ratio

##################### Initialization #####################
cakg = CAKG((kt, kh, kw), G, D, r, n)              
w_s = nn.Parameter(C, C // G, kt, kh, kw)          
unfold = Unfold3d((kt, kh, kw), dilation=1, padding=(kt, kh, kw)//2, stride=1) 
##################### Forward Pass ######################
w_d = cakg(x).view(B, T, G, C // G, C // G, 1, 1, kt, kh, kw)   # Dynamic kernels
w_s = w_s.view(1, 1, G, C // G, C // G, 1, 1, kt, kh, kw)       # Static kernel
x_unfolded = unfold(x).transpose(1, 2)                
x_unfolded = x_unfolded.view(B, T, G, 1, C // G, H, W, kt, kh, kw)
# Apply convolution operations, Eq.(1)
x_d = (w_d * x_unfolded).sum(dim=[4, 7, 8, 9]) # Dynamic   
x_s = (w_s * x_unfolded).sum(dim=[4, 7, 8, 9])  # Static 
out = (x_d + x_s).view(B, T, C, H, W).transpose(1, 2) 
return out
\end{lstlisting}
\vskip -0.075in
\end{algorithm}
\vspace{-0.5 pt}

\subsection{Context-Aware Dynamic Convolution}

Motivated by the demand for dynamic modeling in CSLR and the shortcomings of existing dynamic convolutions in temporal modeling and context awareness—factors that limit their effectiveness in CSLR—we propose the Dynamic Context-Aware Convolution (DCAC). DCAC retains the key advantages of conventional dynamic convolutions while introducing strong context-awareness. In this section, we first present the dual-branch architecture of DCAC, followed by a detailed description of the Context-Aware Kernel Generator (CAKG).

\textbf{Dual-branch structure.}   As shown in the top-left of Fig. \ref{fig:img2}, DCAC consists of a static branch and a dynamic branch. The static branch performs standard convolution, while the dynamic branch applies convolutional kernels generated on the fly. Compared to a purely dynamic design, this dual-branch architecture has been shown to offer greater robustness to noise \cite{bivolution}. Our primary innovation lies in the design of the dynamic branch, and we first present the overall computational formulation of DCAC:
\begin{equation}\label{eq:1} 
  \mathrm{DCAC}(\mathbf{X}) = \underbrace{\mathbf{W^{CAKG}} * \mathbf{X}}_{\text{Dynamic Branch}} + \underbrace{\mathbf{W} * \mathbf{X}}_{\text{Static Branch}},
\end{equation}
where $*$ indicates the convolution operation. $\mathbf{X}\in\mathcal{R}^{C_{i} \times T\times H\times W}$  denotes the input feature with $C_{i}$  represents the number of input channels. $\mathbf{W} \in \mathcal{R}^{C_{o} \times C_{i}/G \times k_{t} \times k_{h} \times k_{w}}$ denotes the static convolutional kernel with a size of $k_t \times k_h \times k_w$, where $C_o$ is the number of output channels and $G$ is the group size. Additionally, $\mathbf{W^{CAKG}}\in\mathcal{R}^{T \times C_{o} \times C_{i}/G \times k_{t}\times k_{h}\times k_{w}}$ represents dynamic weights generated by CAKG. In the formula, we omit the Unfold operation, as it is implicitly included in the convolution operation to ensure parallel computation. In Alg. \ref{alg:code}, we provide the overall computation flow of DCAC in PyTorch-like pseudocode.

\textbf{Context-Aware Kernel Generator.} CAKG introduces a novel paradigm for convolutional kernel generation. As illustrated in Fig.~\ref{fig:img3}, it comprises two main components: Intra-frame Attention and Inter-frame Context Awareness. The former, inspired by ODConv \cite{odconv}, utilizes intra-frame features to generate multi-dimensional attention, while the latter uniquely incorporates contextual information to initialize the kernel weights. The final convolutional weights are obtained by element-wise multiplication of the outputs from these two components. We detail each component below:

\begin{figure}[t!]
  \centering
  \includegraphics[width=\linewidth]{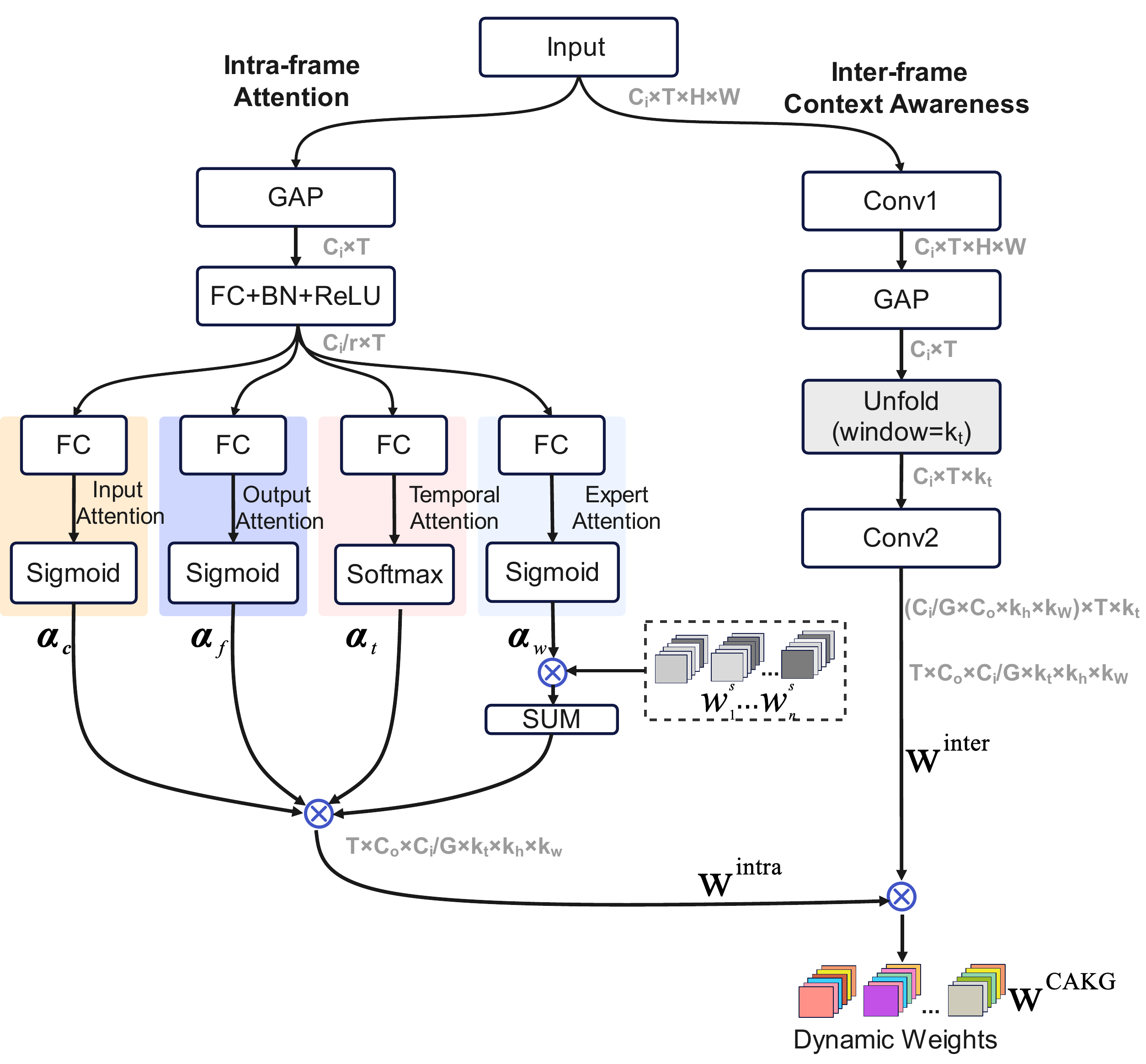}
  \caption{The architecture of the Context-Aware Kernel Generator (CAKG), as defined by Equations \ref{eq:2}--\ref{eq:5}.}
 \label{fig:img3}
\end{figure}

(1) Intra-frame Attention. Attention mechanisms have been proven to play a crucial role in dynamic convolution \cite{yang2019condconv,chen2020dynamic,odconv}. ODConv \cite{odconv} introduces an attention-based paradigm to enhance the effectiveness of dynamic convolution. Inspired by this, we design the Intra-frame Attention module. By leveraging multi-dimensional attention generated from intra-frame features, the dynamic convolution can adaptively adjust its kernels based on frame-specific information, leading to improved performance. The Intra-frame Attention weight $\mathbf{W^{intra}}\in\mathcal{R}^{C_{o} \times C_{i}/G \times T\times k_t\times k_h\times k_w}$  is computed as follows:
\begin{equation}\label{eq:2} 
\mathbf{W^{intra}}=\alpha_{f}\odot\alpha_{c}\odot\alpha_{t}\odot (\alpha_{w1} \odot W_1^s+...+ \alpha_{wn} \odot W_n^s) ,
\end{equation}
where the term $W_{i}^s\in\mathcal{R}^{C_{o} \times C_{i}/G \times T\times k_t\times k_h\times k_w}$ represents the $i$-th  expert (i.e., randomly initialized weight), with a total of $n$ experts. To reduce the number of parameters, all frames share these 
$n$ experts. The attention scalar $\alpha_{wi}\in\mathcal{R}^{T}$ is used to weight $W_{i}^s$ in each frame. Additionally, $\alpha_{f} \in \mathcal{R}^{T \times C_{i}}$ and $\alpha_{c} \in \mathcal{R}^{T \times C_{o}}$ apply attention to the input and output channel of each frame’s kernel, respectively, while $\alpha_{t} \in \mathcal{R}^{T \times k_t}$ modulates the temporal dimension. The operator $\odot$ denotes element-wise multiplication.

As shown in Fig. \ref{fig:img3}, the weights $\alpha_{f}, \alpha_{c}, \alpha_{t}, \alpha_{w}$ across these four dimensions are generated by a multi-head SE-type \cite{senet} mechanism. Specifically, the spatial dimensions of the features are first subjected to global average pooling, followed by a shared fc layer, BatchNorm, and ReLU, yielding $\mathbf{X^{\prime}}\in\mathcal{R}^{T \times C_{i}/r}$, where $r$ is the reduction ratio. Subsequently, $\mathbf{X^{\prime}}$ is processed through four distinct fc layers, each followed by an activation function (either Sigmoid or Softmax), mapping it to $\alpha_{f}, \alpha_{c}, \alpha_{t}, \alpha_{w}$:
\begin{equation}\label{eq:3} 
\begin{aligned}
&\mathbf{X^{\prime}} = \mathrm{ReLU}(\mathrm{BN}(f^{C\to C/r}(\mathrm{GAP}^{H, W\to 1,1}(\mathbf{X})))), \\
&\alpha_{c} = \mathrm{Sigmoid}(f^{C/r \to C_{i}/G}(\mathbf{X^{\prime}})), \\
&\alpha_{f} = \mathrm{Sigmoid}(f^{C/r \to C_{o}}(\mathbf{X^{\prime}})), \\
&\alpha_{t} = \mathrm{Sigmoid}(f^{C/r \to k_{t}}(\mathbf{X^{\prime}})), \\
&\alpha_{w} = \mathrm{Softmax}(f^{C/r \to n}(\mathbf{X^{\prime}})), \\
\end{aligned}
\end{equation}
here, $\mathrm{GAP}^{H , W\to 1 , 1}(\cdot)$ denotes global average pooling that reduces the spatial dimensions $H \times W$ to $1 \times 1$. The operation $f^{a \to b}(\cdot)$ represents a fc layer that maps the input from dimension $a$ to $b$.

(2) Inter-frame Context Awareness. As a visual language, sign language exhibits strong contextual dependencies. In sign language videos, semantic units are conveyed not through isolated frames but through complete temporal context. To fully leverage contextual information and generate context-aware convolutional kernels, we adopt a sliding window mechanism to construct a context space for each frame. This context is then used to initialize the temporal dimension of the frame-specific convolution. Specifically, given the input feature $\mathbf{X}$, the weights are computed as follows:
\begin{equation}\label{eq:4} 
\begin{aligned}
&\mathbf{X^{\prime\prime}} = \mathrm{Unfold}(\mathrm{GAP}^{H, W\to 1,1}(\mathrm{Conv1D}^{C_{i}\to C_{i}}(\mathbf{X}))) ,\\
&\mathbf{W^{inter}} =\mathrm{Reshape}(\mathrm{Conv1D}^{C_{i}\to C_{o} \times C_{i}/G \times k_h \times k_w}(\mathbf{X^{\prime\prime}})), \\
\end{aligned}
\end{equation}
where $\mathrm{Conv1D}^{a \to b}(\cdot)$ denotes a 1D convolution with input and output channels $a$ and $b$, respectively. $\mathrm{Unfold}(\cdot)$ refers to the expansion of features along the temporal axis to construct the temporal dimension of convolutional kernels, using a window size of $k_t$. The resulting intermediate feature is denoted as $\mathbf{X^{\prime\prime}} \in \mathcal{R}^{C_{i} \times T \times k_t}$, and the generated weights are represented by $\mathbf{W^{inter}} \in \mathcal{R}^{C_{o} \times C_{i}/G \times T \times k_t \times k_h \times k_w}$.

Building on this,  the final weight $\mathbf{W^{CAKG}}\in\mathcal{R}^{C_{o} \times C_{i}/G \times T\times k_{t}\times k_{h}\times k_{w}}$ is obtained by multiplying $\mathbf{W^{inter}}$ with $\mathbf{W^{intra}}$, enabling interaction between intra-frame and inter-frame contextual information while reparameterizing the weights jointly:
\begin{equation}\label{eq:5} 
\begin{aligned}
&\mathbf{W^{CAKG}} = \mathbf{W^{inter}} \odot \mathbf{W^{intra}} ,\\
\end{aligned}
\end{equation}
where the  $\mathbf{W^{CAKG}}$ consists of 
$T$ dynamic kernels, each corresponding to a frame of the input feature $\mathbf{X}$. These kernels with a static kernel $\mathbf{W}$ are used to compute the final output of the DCAC following Eq. \ref{eq:1}. Finally, we integrate the output of DCAC into each stage using the approach:
 \begin{equation}\label{eq:6} 
 \mathbf{X}=\mathbf{X}+\alpha \odot \mathrm{DCAC}(\mathbf{X}),
\end{equation}
where $\alpha$ is a learnable parameter that regulates the contribution of DCAC, it is initialized to zero to ensure the preservation of the original features.

\begin{figure}[t!]
  \centering
  \includegraphics[width=\linewidth]{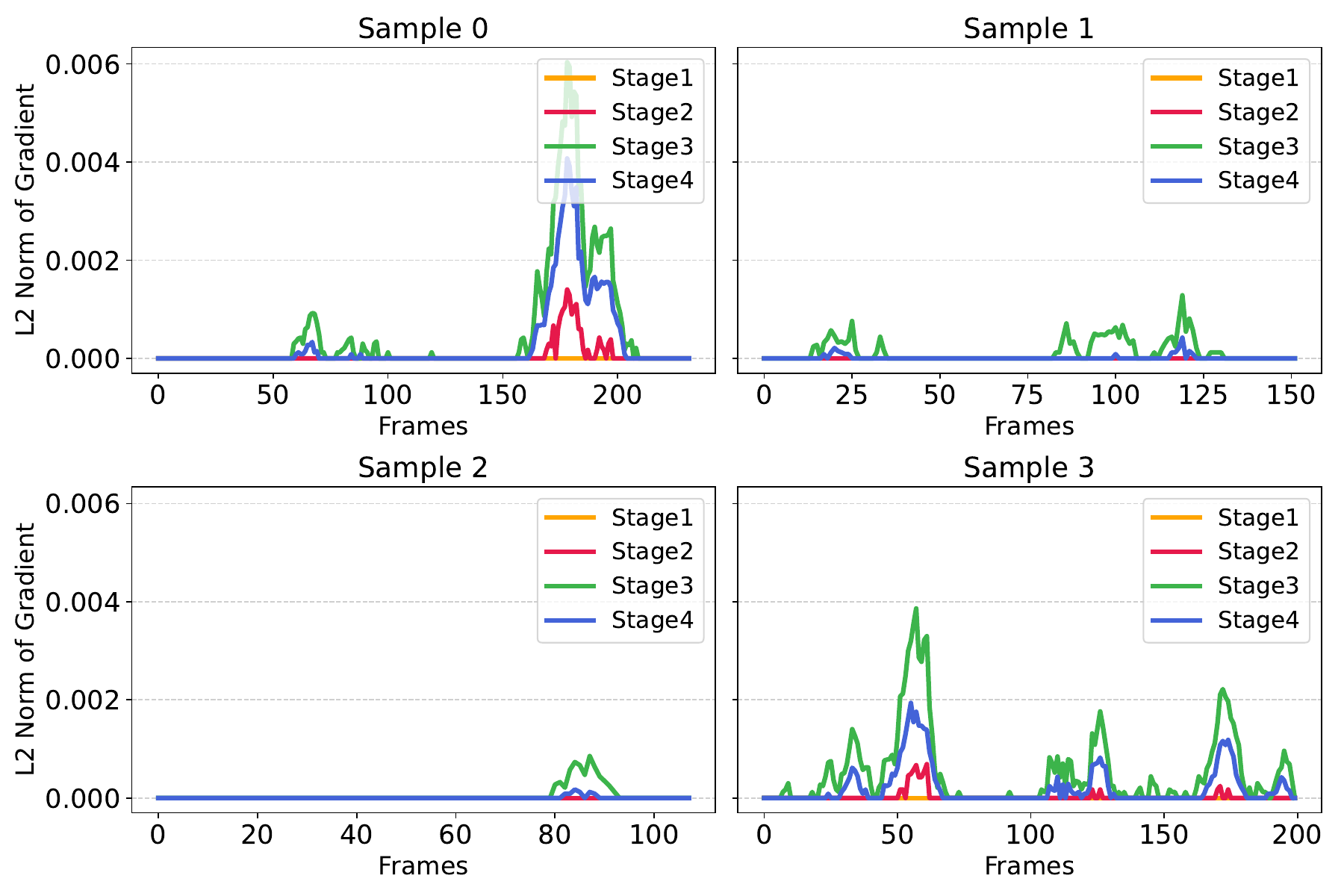}
  \caption{Visualization of frame-wise L2 norm of gradients across different stages for four samples during the later training phase of the baseline model (VAC \cite{vac}+ SMKD \cite{smkd} + TLP \cite{tlp}). Two common patterns can be observed: (1) The gradient exhibits a spike distribution along the frames, indicating that an overfitted model on the dominant path can only update parameters using a small subset of frames. (2) severe gradient vanishing occurs in shallow layers, with stage 1 and stage 2 receiving almost no gradient signal.}
 \label{fig:img4}
\end{figure}

\subsection{SR-CTC}

Although existing CSLR methods \cite{smkd, vac} adopt various strategies to alleviate the peaky nature of CTC, we still observe that gradient signals remain highly concentrated on a small subset of frames during training (as shown in Fig. \ref{fig:img4}). This indicates that the model overly relies on a few frames, which intensifies overfitting along the dominant CTC path. Furthermore, severe gradient vanishing occurs in the shallow layers—particularly in stages 1 and 2—where gradients are almost entirely absent. As a result, parameter updates stagnate in these layers, limiting their ability to learn low-level visual features.

Our SR-CTC addresses the above issue in a simple yet effective manner. As illustrated in Fig.~\ref{fig:img2}, SR-CTC applies auxiliary CTC supervision to the outputs of shallow stages during training.  The core idea is that shallow stages consistently generate low-level features, where gradient signals are more evenly spread across multiple alignment paths, rather than being dominated by a single one. While the overall network still produces peaky predictions, these auxiliary losses act as regularizers, encouraging the model to explore a broader range of alignment possibilities. Additionally, this design provides more direct supervision to earlier layers, helping to alleviate gradient vanishing from deeper layers. To further enhance cross-scale alignment, we introduce a classifier-sharing strategy that promotes consistency among multi-scale features. Owing to its simplicity and modularity, SR-CTC can be seamlessly integrated into existing CSLR pipelines, offering performance improvements with minimal implementation overhead.

\begin{figure}[t!]
  \centering
  \includegraphics[width=\linewidth]{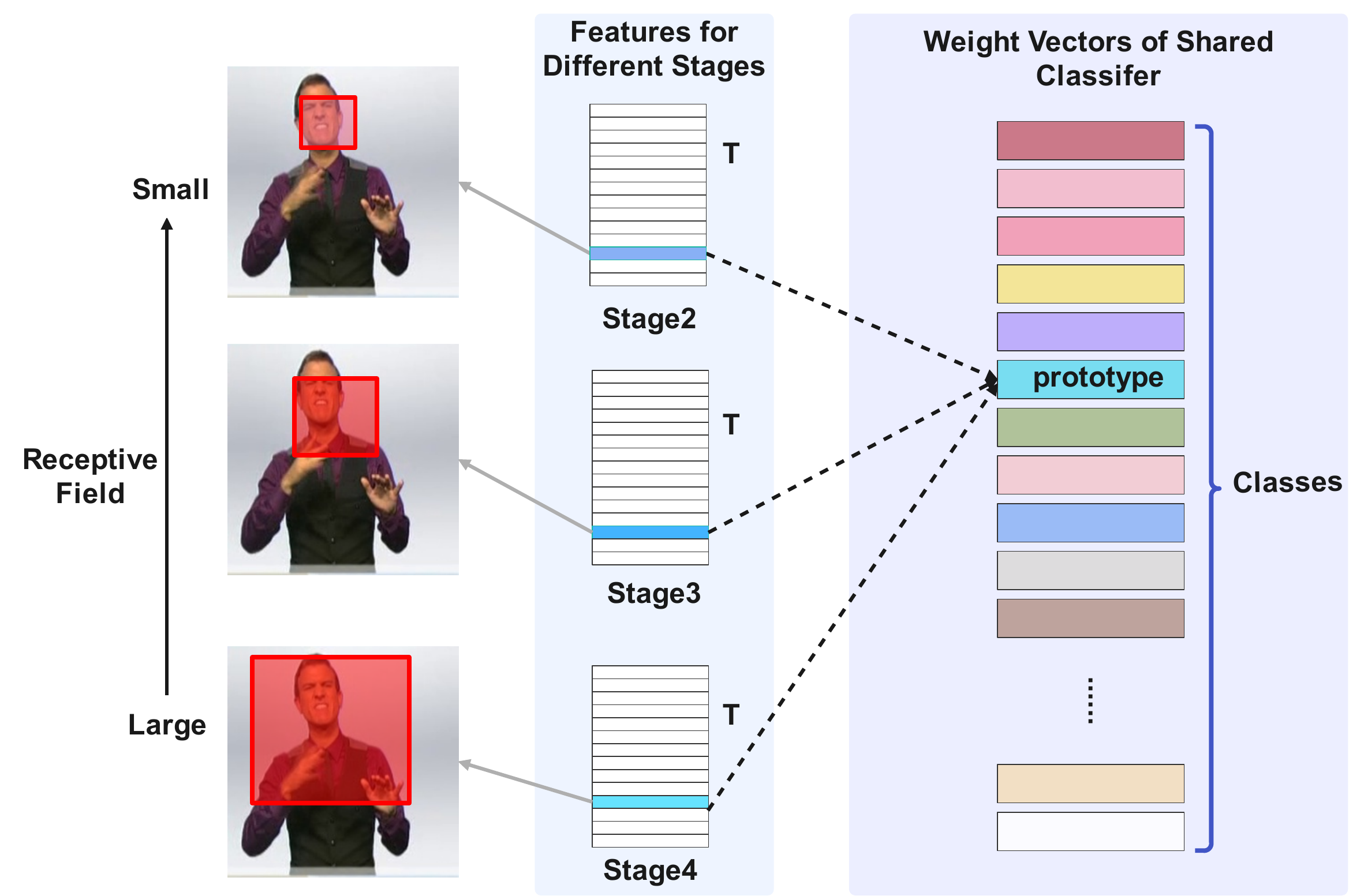}
  \caption{Illustration of classifier-sharing strategy in SR-CTC. This design promotes implicit alignment across scales in the semantic space through shared classifier weights. The red regions highlight the receptive fields corresponding to each stage, and the classifier's weights can be interpreted as class-specific prototypes.}
 \label{fig:img5}
\end{figure}

To enable the effective implementation of SR-CTC, it is crucial to harmonize feature maps across different stages. Since these outputs vary in spatial resolution, channel dimensions, and require temporal alignment, we introduce two lightweight yet effective modules to unify the representations. The details are as follows:

\textbf{Light-weight Spatial Downsampling.} The Light-weight Spatial Downsampling module employs a minimalist design, consisting of a single 2D convolutional layer followed by global average pooling (GAP). Given the input
$\mathbf{X_{i}} \in \mathcal{R}^{C_{i} \times T \times H \times W}$
from the $\mathbf{i}$-th stage, a 2D convolution is first applied to reduce the spatial dimensions to 7×7 while projecting the channel dimension to 512. GAP is then used to further compress the spatial dimensions:
\begin{equation}
\label{eq:7}
\begin{aligned}
 & \mathbf{X^{lsd}_{i}} = \mathrm{GAP}^{7,7\to 1,1}(\mathrm{Conv2D}^{C_{i}\to 512}(\mathbf{X_{i}})),
\end{aligned}
\end{equation}
where $\mathbf{X^{lsd}_{i}} \in \mathcal{R}^{T \times 512}$ denotes the output features. We set the kernel size and stride of the Conv2D layer to $(H//7, W//7)$ to ensure that the resulting spatial dimensions are $7 \times 7$, matching the output shape of the final stage. Consequently, the Conv2D layer in the last stage can be omitted.

\textbf{Light-weight Temporal Modeling.} This module primarily comprises 1D convolution and pooling layers, arranged in the sequence $\{K5, P2, K5, P2\}$, where $K_\sigma$ and $P_\sigma$ denote 1D convolution and pooling layers with a kernel size of $\sigma$, respectively. The operation is defined as:
\begin{equation}
\label{eq:8}
\begin{aligned}
 & \mathbf{X^{ltm}_{i}} = \mathrm{LTM}_{\{K5,P2,K5,P2\}}(\mathbf{X^{lsd}_{i}}) ,
\end{aligned}
\end{equation}
here, $\mathbf{X^{ltm}_{i}} \in \mathcal{R}^{T' \times 1024}$ represents the output. In practical implementation, we found that shared light-weight temporal modeling across stages can achieve better performance, and potentially collaborating with shared classifiers may assist in cross-scale information alignment.
Subsequently, $\mathbf{X^{ltm}_{i}}$ is classified through a classifier:
\begin{equation}
\label{eq:9}
\begin{aligned}
& \mathbf{Z}_i = f_{s}^{1024 \rightarrow Cls} \left( \mathbf{X^{ltm}_{i}} \right).
\end{aligned}
\end{equation}
According to the SMKD \cite{smkd}, the classifier’s weights can be served as classwise prototypes, and the CTC loss drives the model to align each training sample with its corresponding prototype. Here, we employ a shared classifier
\(f_s(\cdot)\) across all stages (Fig.~\ref{fig:img5}), ensuring that multi-scale features are projected into the same prototype-driven space. Consequently, shallow stages receive direct semantic feedback from deeper representations, which enforces consistency across feature scales. Tab. \ref{tab:classifier} provides strong  evidence supporting this claim.

 
For each feature sequence $\mathbf{Z}_i$, the CTC loss \cite{ctc} estimates the probability of the target sequence $\mathcal{G}$ by marginalizing over all valid alignments of length $T'$ that correspond to $\mathcal{G}$.

\begin{equation}
\label{eq:10}
\begin{aligned}
\mathcal{L}_{\mathrm{CTC}}^{(i)} &= -\log p(\mathcal{G} \mid \mathbf{Z}_i) \\
                               &= -\log \left( \sum_{\pi \in \mathcal{B}^{-1}(\mathcal{G})} p(\pi \mid \mathbf{Z}_i) \right),
\end{aligned}
\end{equation}
where $\mathcal{B}^{-1}(\mathcal{G})$ denotes the set of all alignments $\pi$ of length $T'$ that can be collapsed into $\mathcal{G}$ via the CTC mapping, including insertions of the special blank symbol.
The probability of an alignment path $\pi$ is computed under the assumption of conditional independence across time steps:

\begin{equation}
\label{eq:11}
p(\pi \mid \mathbf{Z}_i) = \prod_{t=1}^{T'} p(\pi_t \mid \mathbf{Z}_i),
\end{equation}
where the per-frame probabilities $p(\pi_t \mid \mathbf{Z}_i)$  is obtained via a softmax operation over the outputs.  Finally, the total loss of SR-CTC can be represented as:
\begin{equation}
\label{eq:12}
\mathcal{L}_{\text{SR-CTC}} = \sum_{i \in \mathcal{I}} \mathcal{L}_{\mathrm{CTC}}^{(i)} ,
\end{equation}
where $\mathcal{I}$  represents the set of all stages at which supervision is applied.

\subsection{Total Loss Design}

The total loss of \textbf{DESign}  is defined as:
\begin{equation}\label{eq:13} 
\mathcal{L} =
\underbrace{
\mathcal{L}_{\text{CTC}} + \mathcal{L}_{\text{VA}} + \mathcal{L}_{\text{VE}} + C_{u} + C_{p}
}_{\text{Baseline Loss}} 
+ \lambda \mathcal{L}_{\text{SR-CTC}},
\end{equation}
where $ \mathcal{L}_{\text{CTC}} $ denotes the standard CTC loss applied to the final output of the network, $ \mathcal{L}_{\text{VA}} $ and $ \mathcal{L}_{\text{VE}} $ are the VA and VE losses from the VAC \cite{vac}, and $ C_{u} $ and $ C_{p} $ are the losses from the TLP \cite{tlp}. These components together constitute the \textbf{baseline loss}, and we follow the original configurations from prior work when applying them. 
$ \mathcal{L}_{\text{SR-CTC}} $ represents our proposed \textbf{SR-CTC loss}, and $ \lambda $ is a hyperparameter that balances its contribution. We empirically set $ \lambda = 0.1 $ in the implementation.

\begin{table*}[t!]
    \centering
    \caption{Quantitative comparison with state-of-the-art methods on PHOENIX14 and PHOENIX14-T benchmarks. Asterisk (*) denotes models utilizing feature extractors pretrained on auxiliary video datasets. Abbreviations: of – optical flow, kp – keypoints. }
    \resizebox{\textwidth}{!}{
    \begin{tabular}{llllllccc}
        \toprule
        \multirow{3}{*}{\textbf{Methods}} & \multicolumn{2}{l}{\multirow{3}{*}{\textbf{Cues}}} & \multicolumn{4}{c}{\textbf{PHOENIX14}} & \multicolumn{2}{c}{\textbf{PHOENIX14-T}} \\
        & & & \multicolumn{2}{c}{\textbf{Dev (\%)}} & \multicolumn{2}{c}{\textbf{Test (\%)}} & \textbf{Dev (\%)} & \textbf{Test (\%)} \\
        & & & del/ins & WER $\downarrow$ & del/ins & WER $\downarrow$ & WER $\downarrow$ & WER $\downarrow$ \\
        \midrule
        \multicolumn{9}{c}{\textbf{Video-Only}} \\
        SubUNet (2017)~\cite{subunet} & \multicolumn{2}{l}{video} & 14.6 / 4.0 & 40.8 & 14.3 / 4.0 & 40.7 & - & - \\
        VAC (2021)~\cite{vac} & \multicolumn{2}{l}{video} & 7.9 / 2.5 & 21.2 & 8.4 / 2.6 & 22.3 & 21.4 & 23.9 \\
        SMKD (2021)~\cite{smkd} & \multicolumn{2}{l}{video} & 6.8 / 2.5 & 20.8 & 6.3 / 2.3 & 21.0 & 20.8 & 22.4 \\
        TLP (2022)~\cite{tlp} & \multicolumn{2}{l}{video} & 6.3 / 2.8 & 19.7 & 6.1 / 2.9 & 20.8 & 19.4 & 21.2 \\
        SEN (2023)~\cite{sen} & \multicolumn{2}{l}{video} & 5.8 / 2.6 & 19.5 & 7.3 / 4.0 & 21.0 & 19.3 & 20.7 \\
        AdaBrowse+ (2023)~\cite{Hu2023AdaBrowseAV} & \multicolumn{2}{l}{video} & 6.0 / 2.5 & 19.6 & 5.9 / 2.6 & 20.7 & 19.5 & 20.6 \\
        CorrNet (2023)~\cite{corrnet} & \multicolumn{2}{l}{video} & 5.6 / 2.8 & 18.8 & 5.7 / 2.3 & 19.4 & 18.9 & 20.5 \\
        SlowFastSign* (2023)~\cite{slowfastsign} & \multicolumn{2}{l}{video} & - & 18.0 & - & 18.3 & 17.7 & 18.7 \\
        SignGraph (2024)~\cite{signgraph} & \multicolumn{2}{l}{video} & 4.9 / 2.0 & 18.2 & 4.9 / 2.0 & 19.1 & 17.8 & 19.1 \\
        TCNet (2024)~\cite{tcnet} & \multicolumn{2}{l}{video} & 5.5 / 2.4 & 18.1 & 5.4 / 2.0 & 18.9 & 18.3 & 19.4 \\
        CorrNet+ (2024)~\cite{corrnet+} & \multicolumn{2}{l}{video} & 5.3 / 2.7 & 18.0 & 5.6 / 2.4 & 18.2 & 17.2 & 19.1 \\
        \midrule
        \multicolumn{9}{c}{\textbf{Additional Cues}} \\
        DNF (2019)~\cite{dnf} & \multicolumn{2}{l}{+of} & 7.3 / 3.3 & 23.1 & 6.7 / 3.3 & 22.9 & 22.7 & 24.0 \\
        C$^2$SLR (2022)~\cite{c2slr} & \multicolumn{2}{l}{+kp} & - & 20.5 & - & 20.4 & 20.2 & 20.4 \\
        CVT-SLR (2023)~\cite{cvtslr} & \multicolumn{2}{l}{+text} & 6.4 / 2.6 & 19.8 & 6.1 / 2.3 & 20.1 & 19.4 & 20.3 \\
        CTCA (2023)~\cite{ctca} & \multicolumn{2}{l}{+text} & 6.2 / 2.9 & 19.5 & 6.1 / 2.6 & 20.1 & 19.3 & 20.5 \\
        TwoStream-SLR* (2022)~\cite{twostream} & \multicolumn{2}{l}{+kp} & - & 18.4 & - & 18.8 & 17.7 & 19.3 \\
        SignBERT+* (2023)~\cite{signbert} & \multicolumn{2}{l}{+kp} & 4.8 / 3.7 & 19.9 & 4.2 / 3.8 & 20.0 & 18.8 & 19.9 \\
        C$^2$ST (2023)~\cite{c2st} & \multicolumn{2}{l}{+text} & 4.2 / 3.0 & 17.5 & 4.3 / 3.0& 17.7 & 17.3 & 18.9 \\
        GPGN (2024)~\cite{GPDN} & \multicolumn{2}{l}{+text} & - & 19.9 & - & 20.4 & 19.3 & 20.5 \\
        SignVTCL* (2024)~\cite{signvtcl} & \multicolumn{2}{l}{+of+kp+text} & 6.0 / 2.4 & 17.3 & 5.9 / 2.4 & 17.6 & 16.9 & 17.9 \\
       
        \midrule
        DESign (ours) & \multicolumn{2}{l}{video} & 4.5 / 2.8 &\textbf{ 17.1} & 4.2 / 2.5 & \textbf{17.4} & \textbf{16.5} & \textbf{18.2} \\
        \bottomrule
    \end{tabular}}
    \label{tab:phoenix}
\end{table*}

\section{Experiments}
\label{experiments}
\subsection{Datasets and Evaluation}
In this paper, we primarily evaluate and validate the proposed methods on three large-scale CSLR datasets using the metric of WER. The following provides an overview of the datasets and evaluation metrics:

\textbf{PHOENIX14} \cite{Forster2015Continuous} is a widely used CSLR dataset collected from German TV weather broadcasts performed by nine signers in front of a clean background, with a resolution of 210 × 260. It comprises 6,841 annotated sentences covering 1,295 unique glosses, and is split into 5,672 training samples, 540 development (Dev) samples, and 629 testing (Test) samples.

\textbf{PHOENIX14-T} \cite{8578910} can be serve as an extended version of PHOENIX14, designed to support both CSLR and Sign Language Translation (SLT). It consists of 8,247 annotated sentences and a lexicon of 1,085 unique signs. The dataset is divided into 7,096 training samples, 519 development (Dev) samples, and 642 testing (Test) samples.

\textbf{CSL-Daily} \cite{9578398} is one of the most widely used Chinese sign language datasets for both CSLR and SLT. It covers a broad range of daily-life topics, comprising over 20,000 annotated sentences performed by 10 signers. The dataset is divided into 18,401 training samples, 1,077 development (Dev) samples, and 1,176 test (Test) samples, and includes a vocabulary of 2,000 sign glosses and 2,343 corresponding Chinese words.

\textbf{Evaluation Metric.} Following previous works\cite{vac,smkd,corrnet,tcnet,signgraph}, we adopt Word Error Rate (WER) as the primary evaluation metric to assess model performance. WER quantifies the minimum number of substitutions ($\#$sub), insertions ($\#$ins), and deletions ($\#$del) needed to align the predicted sequence with the ground-truth reference ($\#$ref), and is defined as:
\begin{equation}\label{eq:14} \mathrm{WER}=\frac{\#\mathrm{sub}+\#\mathrm{ins}+\#\mathrm{del}}{\#\mathrm{ref}}. \end{equation}
Note that a \textbf{lower} WER corresponds to \textbf{higher} accuracy.

\textbf{Network Details.}
 For the feature extractor, we employ a ResNet34 \cite{resnet} pretrained on ImageNet \cite{imagenet}.
 The 1D-CNN module comprises two temporal 1D convolutional layers, each followed by an adaptive temporal downsampling layer \cite{tlp}. This is followed by a two-layer Bi-LSTM encoder with 1,024 hidden units, and a fully connected layer for gloss prediction. For DCAC, we ultimately adopted a depthwise separable convolution approach and set $k_h$ and $k_w$ to 1 to reduce computational complexity.
Based on previous works \cite{yang2019condconv,odconv}, we empirically determined the number of experts $n$ to be 6 (Eq. \ref{eq:2}), and set the reduction ratio 
$r$ to 16 (Eq. \ref{eq:3}).

\textbf{Training and Testing Strategy.}
 During training, we set the batch size to 2 and initialize the learning rate at 0.001, reducing it to 20\% of its previous value at epochs 40 and 60. We use the Adam optimizer with a weight decay of 0.001 and train for a total of 80 epochs. After each epoch, we compute the WER on the Dev set and save checkpoints corresponding to the lowest WER observed.
 All input frames are first resized to 256×256, then randomly cropped to 224×224 during training, with a 50\% chance of horizontal flipping and a ±20\% probability of temporal scaling. For inference, we only use a center crop of 224×224 and apply a beam search algorithm with a beam width of 10 for CTC decoding. Our SR-CTC auxiliary modules, LST and LTM, are only activated during the training phase. Finally, all training and testing are conducted on a single NVIDIA A6000 GPU.
\begin{table}[t!]
 \caption{Quantitative comparison with state-of-the-art methods on CSL-Daily. Asterisk (*) denotes models utilizing feature extractors pretrained on auxiliary video datasets. Abbreviations: of – optical flow, kp – keypoints.}
\centering 
   \resizebox{0.5\textwidth}{!}{
    \begin{tabular}{llcc}
    \toprule
    \textbf{Methods} &\textbf{Cues} &\textbf{Dev (\%)} & \textbf{Test (\%)} \\ 
    \midrule 
    \multicolumn{4}{c}{\textbf{Video-Only}}\\
    LS-HAN (2018)~\cite{lshan} &video &39.0 & 39.4 \\
    SEN (2023)~\cite{sen} &video &31.1&30.7\\
    AdaBrowse+ (2023)~\cite{Hu2023AdaBrowseAV} &video &31.2&30.7\\
    CorrNet (2023)~\cite{corrnet} &video &30.6 & 30.1 \\
    SlowFastSign* (2023)~\cite{slowfastsign} &video &25.5 & 24.9 \\
    TCNet (2024)~\cite{tcnet} &video &29.7 & 29.3 \\
    CorrNet+ (2024)~\cite{corrnet+} &video &28.6 & 28.2 \\
    SignGraph (2024)~\cite{signgraph} &video &27.3 & 26.4 \\
    Swin-MSTP (2025)~\cite{alyami2025swin} &video &28.3 & 27.1 \\
    \midrule
    \multicolumn{4}{c}{\textbf{Additional Cues}}\\
    TwoStream-SLR* (2022)~\cite{twostream} &+kp &25.4 &25.3\\ 
    CTCA (2023)~\cite{ctca} &+text &31.3 &29.4\\ 
    C$^2$ST (2023)~\cite{c2st} &+text&25.9 &25.8\\ 
    GPGN (2024)~\cite{GPDN} &+text &31.1 &30.0\\ 
    SignVTCL* (2024)~\cite{signvtcl} &+of+kp+text&24.3 &24.1\\ 
    Uni-Sign* (2025)~\cite{unisign} &+kp+text&26.7 &26.0\\ 
    \midrule 
    DESign (ours) &video&\textbf{25.3}& \textbf{24.2}\\
    \bottomrule
    \end{tabular} }
\label{tab:csl}
 \vspace{-\baselineskip}
\end{table}

\vspace{-0.5 pt}

\vspace{-0.5 pt}

\subsection{Comparison with state-of-the-art Methods}
We evaluate the performance of DESign on the three datasets and compare it with state-of-the-art methods. Additionally, we present separate comparisons of DCAC and SR-CTC against their related methods in the \textbf{supplementary materials} to highlight their individual contributions.

\textbf{On PHOENIX14 and PHOENIX14-T.} As shown in Tab. \ref{tab:phoenix}, we compare DESign with existing state-of-the-art methods on PHOENIX14 and PHOENIX14-T. For a fair comparison, all methods are categorized into single-cue (Video-Only) and multi-cue (Additional Cues), where multi-cue methods leverage additional modalities (e.g., optical flow, text, keypoints) on top of RGB inputs to improve performance. We also indicate methods that leverage additional pretraining on external video datasets (e.g., Kinetics).

DESign, as a single-cue model, achieves state-of-the-art performance without relying on any auxiliary video datasets. For example, SlowFastSign \cite{slowfastsign} employs a large-scale feature extractor, SlowFast-101 \cite{slowfast}, pretrained on an action recognition dataset to obtain robust visual features. In contrast, benefiting from the proposed DCAC and SR-CTC modules, DESign employs only a lightweight ResNet34 \cite{resnet}, yet achieves an absolute WER reduction of 0.9\% on both the Dev and Test sets of PHOENIX14. Moreover, it yields further absolute reductions of 1.2\% Dev set and 0.5\% on the Test set of PHOENIX14-T.

It's also worth noting that the multi-cue state-of-the-art method SignVTCL \cite{signvtcl} adopts an extremely complex architecture. It employs an S3D \cite{xie2018rethinking} model pretrained on Kinetics-400 as the feature extractor and constructs a sophisticated three-stream framework that fuses RGB frames, optical flow, and keypoint information, further enhanced by auxiliary supervision from textual inputs. Nevertheless, DESign, with its low-latency and easily deployable architecture, achieves absolute WER reductions of both 0.2\% on the Dev and Test sets of PHOENIX14, surpassing SignVTCL despite its simpler design. As a result, DESign not only achieves competitive performance but also offers a more practical and efficient solution for real-world CSLR deployment.

\textbf{On CSL-Daily.} According to Tab. \ref{tab:csl}, DESign also demonstrates strong competitiveness on a large-scale Chinese dataset. It consistently outperforms all single-cue methods, including SlowFastSign \cite{slowfastsign}, which benefits from pretraining on additional video data. In the comparison with multi-cue models, DESign surpasses TwoStream \cite{twostream} by 0.1\% (Dev) and 1.1\% (Test), despite the latter leveraging a large pretrained feature extractor and pre-extracted keypoints to focus on salient video regions. Moreover, although Uni-Sign \cite{unisign} is pretrained on a self-constructed large-scale sign language dataset and incorporates keypoint information to boost performance, DESign still achieves a WER reduction of 1.4\% (Dev) and 1.8\% (Test). These results further highlight DESign’s robustness and generalization capability across different languages and sign language datasets.

\begin{table}[t]
\caption{Ablations for the effectiveness of DCAC and SR-CTC on the PHOENIX14}
\centering
\resizebox{0.35\textwidth}{!}{
    
    \begin{tabular}{cccc}
        \toprule
         DCAC & SR-CTC& Dev  (\%)& Test (\%)\\
        \midrule
        \xmark&\xmark& 19.2& 19.5\\
        \cmark & \xmark& 17.9 & 18.0\\
        \xmark&\cmark & 17.8& 18.9\\
        \cmark &\cmark & \textbf{17.1}& \textbf{17.4}\\
        \bottomrule
    \end{tabular}     
   }
    \label{tab:ablation1}
\end{table}

\begin{table}[t]
\caption{Ablations for the main components of DCAC on the PHOENIX14}
\centering
\resizebox{0.45\textwidth}{!}{
    
    \begin{tabular}{c|cc|cc}
        \toprule
         \multirow{2}{*}{Static Branch} & \multicolumn{2}{c|}{Dynamic Branch} &  \multirow{2}{*}{Dev  (\%)}& \multirow{2}{*}{ Test (\%)}\\
         & CAKG (inter) & CAKG (intra) & &\\

        \midrule
        \xmark&\xmark&\xmark& 19.2& 19.5\\
        \cmark & \xmark&\xmark& 18.6 & 19.0\\
        \xmark&\cmark &\xmark& \textbf{17.9}& 18.6\\
         \xmark&\xmark &\cmark& 18.1& 18.7\\
        \cmark &\cmark &\cmark& \textbf{17.9}& \textbf{18.0}\\
        \bottomrule
    \end{tabular}     
   }
    \label{tab:dcac1}
\end{table}

\begin{table}[t]
\centering
\caption{Ablations for temporal receptive field configurations in DCAC. $L = [L_1, L_2, L_3]$ denotes the temporal receptive field in stages 2, 3, and 4, respectively. FLOPs are measured on a 100-frame video.}
\label{tab:dcac2}
\resizebox{0.4\textwidth}{!}{    
    \begin{tabular}{l|ccc}
        \toprule
        Configuration & Dev (\%) & Test (\%)&FLOPs (G) \\
        \midrule
        -     &     19.2 & 19.5  & 369.46  \\
        $L=[3, 3, 3]$     &     18.3 & 19.2  & 373.48   \\
        $L=[3, 5, 7] $    &   18.1       &   18.5    & 373.62   \\
        $L=[3, 7, 11]$    &      17.9  & \textbf{18.0} &         373.75 \\
        $L=[5, 7, 9] $   &  18.0        & 18.3         &   373.72  \\
        $L=[5, 9, 13]$    & \textbf{17.6}         & 18.3      & 373.86    \\
        $L=[13, 13, 13] $ &          17.8 & 18.1       & 374.00    \\
        \bottomrule
    \end{tabular}
}
\end{table}

\begin{table}[t]
\caption{Ablations for the locations of DCAC on the PHOENIX14. Stage n indicates the location after the n-th stage}
\centering
\resizebox{0.43\textwidth}{!}{
    
    \begin{tabular}{ccccc}
        \toprule
         Stage 2&Stage 3& Stage 4& Dev  (\%)& Test (\%)\\
        \midrule
        \xmark&\xmark& \xmark& 19.2& 19.5\\
        \cmark & \xmark& \xmark& 18.8 & 19.1\\
        \xmark&\cmark & \xmark& 18.7& 18.9\\
        \xmark&\xmark & \cmark& 18.4 & 18.6\\
        \xmark & \cmark & \cmark & 18.2 & 18.4\\
        \cmark &\cmark & \cmark & \textbf{17.9}& \textbf{18.0}\\
        \bottomrule
    \end{tabular}     
   }
    \label{tab:dcac3}
\end{table}

\begin{table}[t]
\caption{Ablations for the effect of weight sharing in SR-CTC‘s classifiers on PHOENIX14.  $f_s^{i}$ indicates the auxiliary classifier for the $i$-th stage, and $f^{\mathrm{f}}$ denotes the final classifier responsible for the model's main output}
\centering
\resizebox{\linewidth}{!}{
\begin{tabular}{lccc}
\toprule
Configuration & Dev (\%) & Test (\%) \\
\midrule
- &19.2 &19.5\\
1) Shared: $f_s^{2}$, $f_s^{3}$, $f_s^{4}$; Unshared: $f^{\mathrm{f}}$ &  18.7&19.3  \\
2) All Shared + Frozen: $f_s^{2}$, $f_s^{3}$, $f_s^{4}$;  Unfrozen: $f^{\mathrm{f}}$ &18.9&19.4  \\
3) Shared: $f_s^{2}$, $f_s^{3}$, $f_s^{4}$, $f^{\mathrm{f}}$ & \textbf{17.8 }& \textbf{18.9} \\
4) Unshared: $f_s^{2}$, $f_s^{3}$, $f_s^{4}$, $f^{\mathrm{f}}$ &  18.8&19.5 \\
\bottomrule
\end{tabular}
}
\label{tab:classifier}
\end{table}

\begin{table}[t]
\caption{Ablations for the locations of SR-CTC on the PHOENIX14. Stage n indicates the location after the n-th stage}
\centering
\resizebox{0.45\textwidth}{!}{
    
    \begin{tabular}{ccccc}
        \toprule
         Stage 2&Stage 3& Stage 4& Dev  (\%)& Test (\%)\\
        \midrule
        \xmark&\xmark& \xmark& 19.2& 19.5\\
        \xmark & \xmark& \cmark& 18.8 & 19.3\\
        \xmark&\cmark & \cmark& 18.3& 18.9\\
        \cmark&\cmark & \cmark& \textbf{17.8} & \textbf{18.9}\\
        \bottomrule
    \end{tabular}     
   }
    \label{tab:sgctc}
\end{table}

    

\begin{table}[t]
\centering
\caption{Evaluating the impact of SR-CTC on existing CSLR methods on PHOENIX14, with all other configurations following the original papers and publicly available codes}
\resizebox{0.36\textwidth}{!}{
\begin{tabular}{lcc}
\toprule
\textbf{Configuration} & \textbf{Dev (\%)} & \textbf{Test (\%)} \\
\midrule
SEN \cite{sen}      & 19.5  &21.0 \\
\quad w/ SR-CTC     &19.1($\downarrow$ 0.4)   &19.1 ($\downarrow$ 1.9)  \\
\midrule
TLP \cite{tlp}      & 19.7  &20.8 \\
\quad w/ SR-CTC     &  18.3 ($\downarrow$ 1.4) & 18.7 ($\downarrow 2.1$) \\
\midrule
CorrNet \cite{corrnet}   &18.9 & 19.7   \\
\quad w/ SR-CTC     &  18.6 ($\downarrow$ 0.3) &19.1 ($\downarrow 0.6 $)  \\
\midrule
CorrNet+ \cite{corrnet+}    & 18.0  &18.2  \\
\quad w/ SR-CTC     & 17.7 ($\downarrow$ 0.3)  & 18.0 ($\downarrow$ 0.2) \\

\bottomrule
\label{tab:pap}
\end{tabular}
}
\end{table}
\vspace{-15pt}

\subsection{Ablation Studies}
In this section, we present comprehensive ablations on each component of DESign using the PHOENIX14 dataset, along with detailed experimental results and analyses. Additionally, we validate the generalizability of DESign across various feature extractors in the supplementary materials.

\textbf{Ablations for DCAC and SR-CTC.} Tab. \ref{tab:ablation1} presents the ablation results for DCAC and SR-CTC. It can be observed that using DCAC or SR-CTC independently both yield performance improvements. Specifically, DCAC reduces the WER by 1.3\% (Dev) and 1.5\% (Test) compared to the baseline, while SR-CTC lowers the baseline WER by 1.4\% (Dev) and 0.6\% (Test) without introducing any additional inference cost. When both DCAC and SR-CTC are employed together, the performance peaks, achieving a WER reduction of 2.0\% (Dev) and 1.9\% (Test) relative to the baseline. Ultimately, DESign incorporating both DCAC and SR-CTC yields the best results.

\textbf{Ablations for the main components of DCAC.} 
As shown in Tab. \ref{tab:dcac1}, we conduct an ablation on the main components of DCAC. In the Dynamic Branch, 'CAKG (inter)' and 'CAKG (intra)' refer to Intra-frame Attention and Inter-frame Context Awareness, respectively. The results show that all components contribute to performance improvements over the baseline. Notably, 'CAKG (inter)' yields the most significant gains among all components, reducing WER by 1.3\% (Dev) and 0.9\% (Test), highlighting the importance of context-aware modeling in CSLR. Finally, DCAC achieves the best overall performance when integrating all components from both the Static and Dynamic branches, with absolute WER reductions of 1.3\% (Dev) and 1.5\% (Test), respectively. These findings validate the effectiveness of each design choice within the DCAC module.

\textbf{Ablations for temporal receptive field configurations in DCAC.} 
Tab.~\ref{tab:dcac2} presents an ablation study of the temporal receptive field configurations in DCAC, where $L = [L_1, L_2, L_3]$ denotes the temporal receptive field size $k_t$ in stages 2, 3, and 4, respectively. All tested configurations show significant reductions in WER compared to the baseline. Among them, the configuration $L = [5, 9, 13]$ achieves the lowest WER on the Dev set (17.6\%), while $L = [3, 7, 11]$ yields the best performance on the test set with a WER of 18.0\%. Although larger receptive fields such as $L = [13, 13, 13]$ also perform competitively, they do not offer further improvements and slightly increase the computational cost. Taking both performance and FLOPs into consideration, we adopt $L = [3, 7, 11]$ as the final configuration for DCAC.

\textbf{Ablations for the locations of DCAC.} 
As shown in Tab. \ref{tab:dcac3}, we conduct ablations on the placement of the DCAC module, which is inserted after stages 2, 3, and 4 of the feature extractor (stage 1 is excluded due to its low-level features \cite{corrnet,corrnet+}). The results demonstrate that integrating DCAC at any stage consistently enhances model performance. Notably, the performance gains become more pronounced at later stages, likely due to the larger receptive field and richer semantic representations of the features. Specifically, inserting DCAC after stage 2 yields a WER reduction of 0.4\% on both the Dev and Test sets compared to the baseline. Placing it after stage 4 achieves a more significant improvement, reducing WER by 0.8\% (Dev) and 0.9\% (Test). Combining DCAC at both stages 3 and 4 further reduces WER to 18.2\% (Dev) and 18.4\% (Test). The best performance is achieved when DCAC is placed after stages 2, 3, and 4 simultaneously, resulting in the lowest WER of all settings—a 1.3\% (Dev) and 1.5\% (Test) absolute reduction over the baseline. This configuration is adopted as the default setup for DESign.

\textbf{Ablations for the effect of weight sharing in SR-CTC‘s classifiers.}
As shown in Tab. \ref{tab:classifier}, we investigate the impact of the classifier-sharing strategy in SR-CTC. Specifically, we evaluate four configurations based on whether the classifiers across different stages share weights and whether those weights are frozen during training. All strategies lead to performance improvements over the baseline. Notably, configuration 3 (shares weights across all classifiers without freezing) achieves the best performance. We attribute this to the fact that the final classifier $f^{\mathrm{f}}$ can be viewed as a set of class-wise prototypes that map features into a semantic space. When all stages share the same classifier, multi-scale features are projected into a more consistent space. This enables effective alignment across scales by optimizing the similarity between features and prototypes through the CTC loss \cite{smkd}. Moreover, allowing the shared classifier to be trainable further enhances this alignment and leads to better overall performance. Finally, we adopt configuration 3 as the classifier-sharing strategy for SR-CTC.

\textbf{Ablations for the locations of SR-CTC.}
Tab.~\ref{tab:sgctc} presents the ablation results of applying SR-CTC at different stages, specifically after stage 2, 3, or 4 (We omit stage 1 due to its low-level features). According to the results, applying SR-CTC only after stage 4 yields a WER reduction of 0.3\% (Dev) and 0.2\% (Test) compared to the baseline. Introducing SR-CTC after both stage 3 and stage 4 leads to a further improvement, with absolute WER reductions of 0.9\% (Dev) and 0.6\% (Test). Notably, applying SR-CTC after stage 2, 3, and 4 achieves the best performance, reducing WER by 1.4\% (Dev) and 0.6\% (Test) compared to the baseline. These results suggest that supervising multiple stages with SR-CTC provides better regularization during training and effectively mitigates CTC’s overfitting to dominant paths. By default, our SR-CTC adopts this configuration.

\textbf{Evaluating the impact of SR-CTC on existing CSLR methods.} SR-CTC can be easily integrated into existing CSLR frameworks and improves performance without incurring any additional inference overhead. As shown in Tab. \ref{tab:pap}, SR-CTC significantly enhances the performance of several representative CSLR models. Most notably, it reduces the WER by 1.4\% (Dev) and 2.1\% (Test) on TLP \cite{tlp}. Even for the current single-cue SOTA CorrNet+ \cite{corrnet+}, SR-CTC brings further improvements, reducing the WER by 0.3\% (Dev) and 0.2\% (Test). These results demonstrate the broad compatibility and practical value of SR-CTC for advancing CSLR.

\begin{figure*}
\centering
  \includegraphics[width=0.85\textwidth]{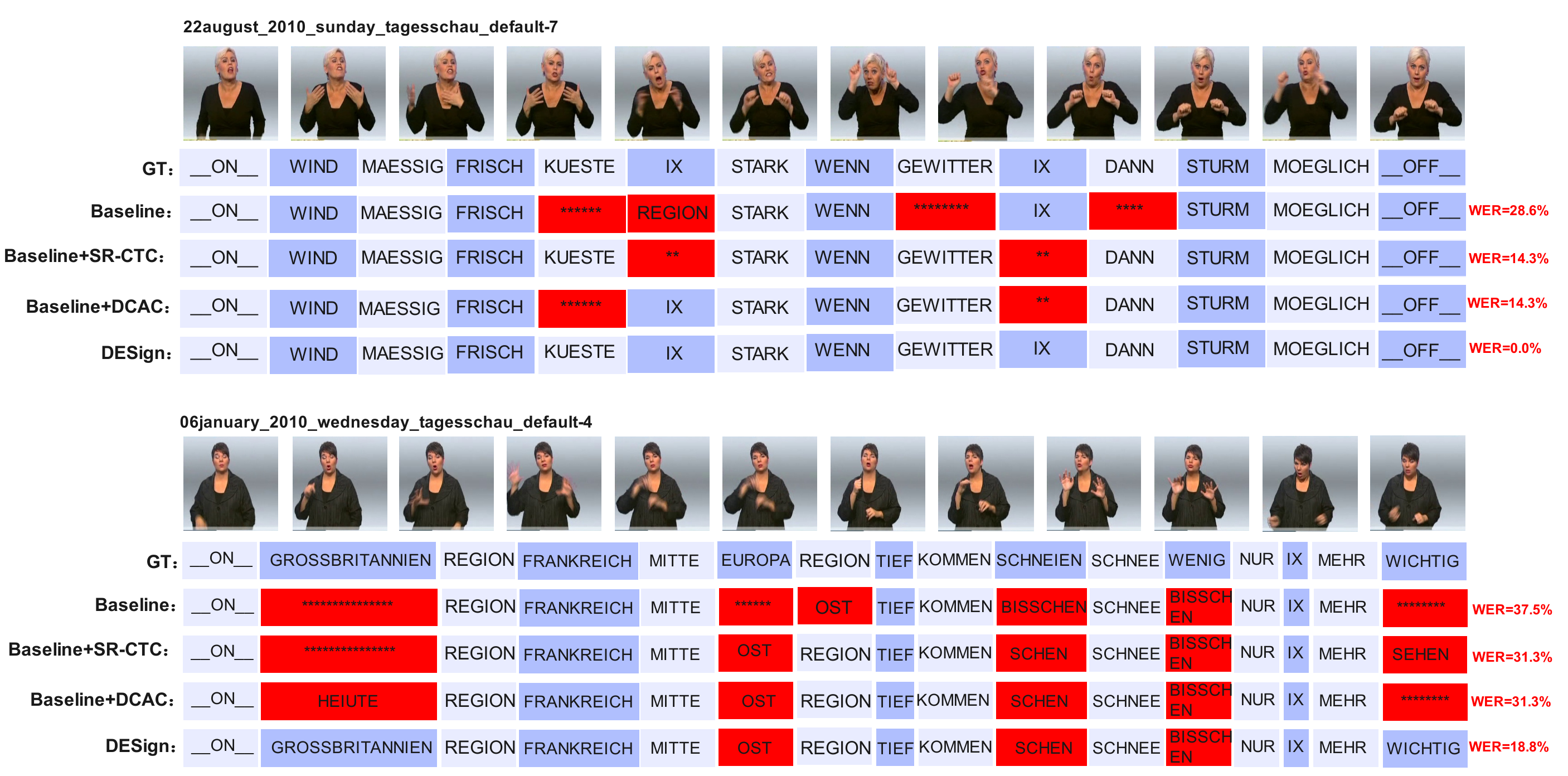}
  \caption{Qualitative comparison of recognition results among Baseline, Baseline+SR-CTC, Baseline+DCAC, and DESign (i.e., Baseline+SR-CTC+DCAC) on sample sequences from the PHOENIX14. Incorrectly predicted glosses are highlighted in red. Word Error Rate (WER) for each method is shown at the end of each hypothesis.}
 
  \label{fig:qualitative}
\end{figure*}

\begin{figure*}
\centering
  \includegraphics[width=0.8\textwidth]{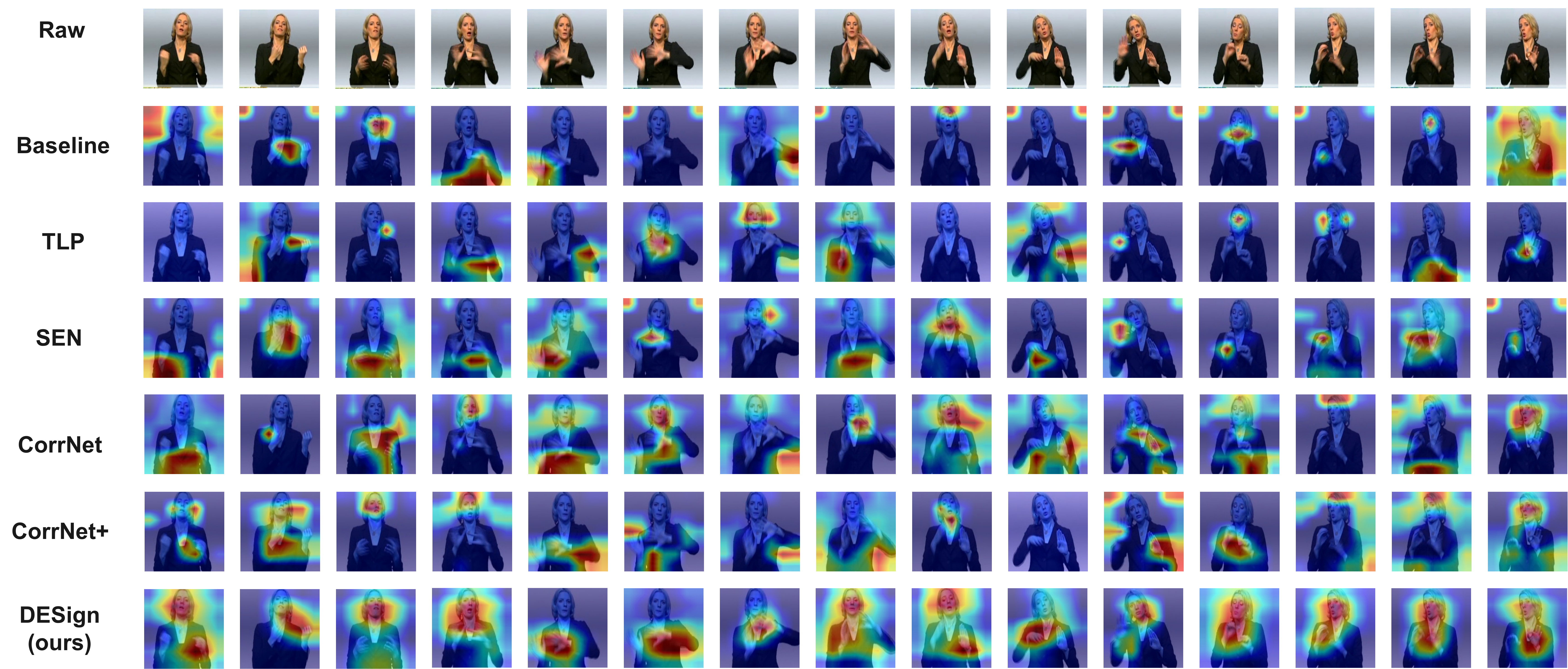}
  \caption{Heatmap comparison between DESign and prior methods (i.e., Baseline, TLP \cite{tlp}, SEN \cite{sen}, CorrNet \cite{corrnet}, and CorrNet+ \cite{corrnet+}). Red areas denote high model attention, while blue indicates low attention. Compared to other methods, DESign more effectively concentrates on informative regions (e.g., hands, face, and body).}
 
  \label{fig:gradcam}
\end{figure*}


\section{Visualizations}
\textbf{Qualitative Comparison of DESign.} To qualitatively analyze the contributions of each component in DESign, we present a comparison of recognition results on two samples in Fig. \ref{fig:qualitative}. In the first example (top row), SR-CTC corrects errors made by the Baseline in recognizing the glosses “KUESTE”, “GEWITTER”, and “DANN”. Additionally, DCAC further rectifies the misclassification of “IX”. Ultimately, DESign successfully recognizes the entire gloss sequence with no errors. In the second example (bottom row), both SR-CTC and DCAC correct the gloss “REGION”, while DESign further corrects “GROSSBRITANNIEN” and “WICHTIG”, achieving the lowest WER among all settings. These results clearly illustrate the complementary strengths of SR-CTC and DCAC, and their combined effectiveness in enhancing performance.

\begin{figure}[t!]
  \centering
  \includegraphics[width=\linewidth]{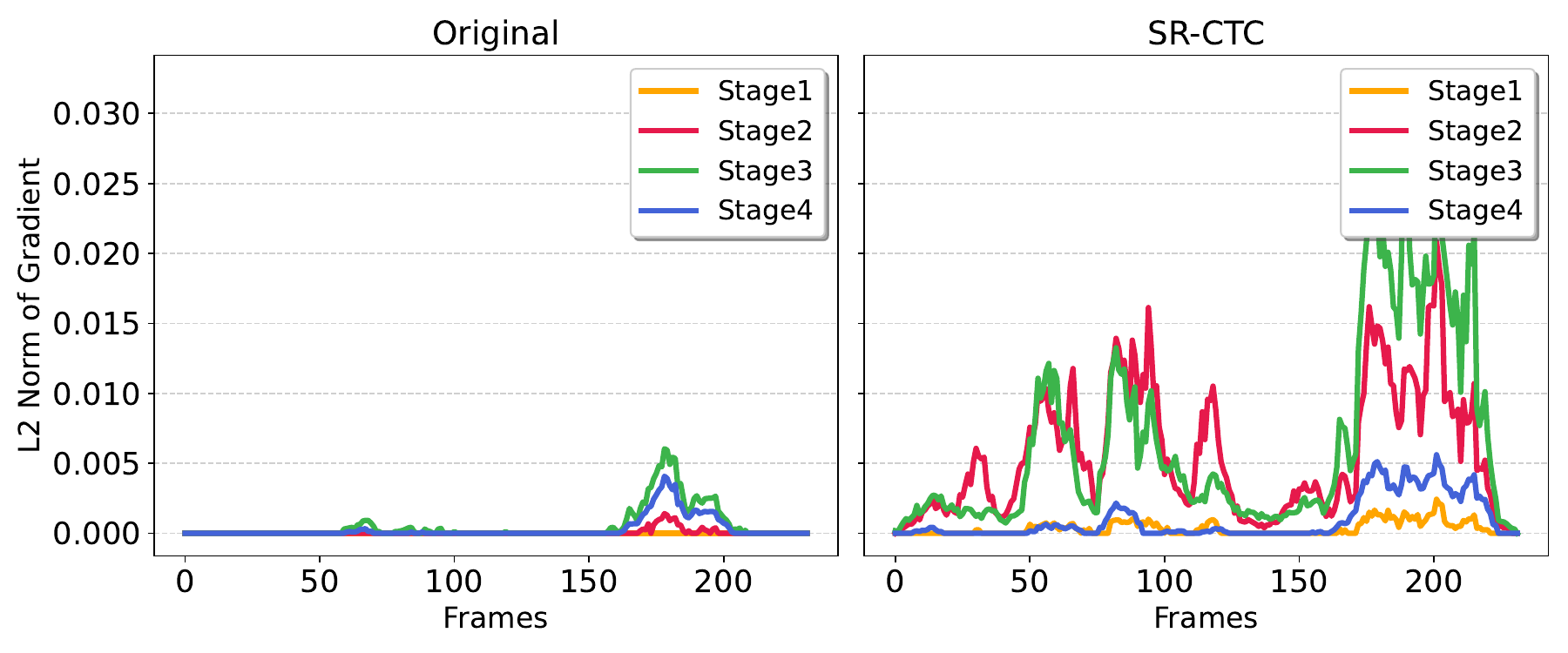}
  \caption{Comparison of frame-wise gradient L2 norms for a video sample in the later training stage, before and after applying SR-CTC. SR-CTC effectively mitigates vanishing gradients and spiky distributions, facilitating more effective utilization of input frames for parameter updates and reducing the risk of overfitting.}
 \label{fig:gl2norm}
\end{figure}

\textbf{Gradients of the frame-wise L2 norm comparison before and after using SR-CTC.}
In Fig. \ref{fig:gl2norm}, we select a sample and track the L2 norm of gradients frame by frame during the later stage of training, comparing the results with and without SR-CTC. Two key observations emerge: (1) With SR-CTC, gradient signals are spread across more frames rather than being concentrated on a few key ones; (2) The gradient magnitudes in shallow stages increase significantly. These findings confirm that SR-CTC effectively regularizes the training process, preventing spiky gradient distributions caused by overfitting to dominant alignment paths. Furthermore, the increased values in shallow layers provide additional evidence that SR-CTC mitigates gradient vanishing. Together, these results demonstrate the role of SR-CTC in enabling more stable and efficient training for CSLR.

\textbf{Heatmap comparison between DESign and previous methods.} The ability to localize informative regions often reflects a CSLR model’s recognition quality. As shown in Fig.~\ref{fig:gradcam}, Grad-CAM \cite{gradcam} visualizations on the PHOENIX14  reveal that DESign focuses more accurately on semantically relevant areas (e.g., hands and face), while prior methods tend to attend to background regions, potentially degrading performance. This highlights DESign’s strength in capturing meaningful visual cues.

\section{Conclusion}

In this paper, we propose DESign, a novel framework for continuous sign language recognition (CSLR) that introduces two key components: Dynamic Context-Aware Convolution (DCAC) and Subnet Regularization CTC (SR-CTC). As a novel dynamic convolution tailored for CSLR, DCAC not only dynamically captures inter-frame motion cues, but also finely adjusts its weights based on contextual information to handle complex temporal variations. SR-CTC introduces a simple yet effective regularization strategy that facilitates better alignment learning by imposing auxiliary supervision on sub-networks. In addition, a classifier-sharing mechanism further promotes implicit alignment consistency across different scales. Notably, SR-CTC is a plug-and-play method that introduces no additional inference overhead. Extensive experiments on PHOENIX14, PHOENIX14-T, and CSL-Daily demonstrate that DESign achieves state-of-the-art performance without any external cues or auxiliary data. We hope DESign will serve as a strong baseline for future CSLR research and inspire further advances in dynamic convolution.

\ifCLASSOPTIONcaptionsoff
  \newpage
\fi

{
\bibliographystyle{IEEEtran}
\bibliography{ieee}

\begin{thebibliography}{10}
\providecommand{\url}[1]{#1}
\csname url@samestyle\endcsname
\providecommand{\newblock}{\relax}
\providecommand{\bibinfo}[2]{#2}
\providecommand{\BIBentrySTDinterwordspacing}{\spaceskip=0pt\relax}
\providecommand{\BIBentryALTinterwordstretchfactor}{4}
\providecommand{\BIBentryALTinterwordspacing}{\spaceskip=\fontdimen2\font plus
\BIBentryALTinterwordstretchfactor\fontdimen3\font minus \fontdimen4\font\relax}
\providecommand{\BIBforeignlanguage}[2]{{%
\expandafter\ifx\csname l@#1\endcsname\relax
\typeout{** WARNING: IEEEtran.bst: No hyphenation pattern has been}%
\typeout{** loaded for the language `#1'. Using the pattern for}%
\typeout{** the default language instead.}%
\else
\language=\csname l@#1\endcsname
\fi
#2}}
\providecommand{\BIBdecl}{\relax}
\BIBdecl

\bibitem{dreuw2007speech}
P.~Dreuw, D.~Rybach, T.~Deselaers, M.~Zahedi, and H.~Ney, ``Speech recognition techniques for a sign language recognition system,'' \emph{hand}, vol.~60, no.~80, 2007.

\bibitem{ong2005automatic}
S.~C. Ong and S.~Ranganath, ``Automatic sign language analysis: A survey and the future beyond lexical meaning,'' \emph{IEEE Transactions on Pattern Analysis \& Machine Intelligence}, vol.~27, no.~06, pp. 873--891, 2005.

\bibitem{Reagan_2007}
T.~Reagan, ``Sign language and linguistic universals, wendy sandler and diane lillo-martin,'' \emph{Studies in Second Language Acquisition}, vol.~29, no.~4, p. 624–625, 2007.

\bibitem{alyami2024reviewing}
S.~Alyami, H.~Luqman, and M.~Hammoudeh, ``Reviewing 25 years of continuous sign language recognition research: Advances, challenges, and prospects,'' \emph{Information Processing \& Management}, vol.~61, no.~5, p. 103774, 2024.

\bibitem{freeman1995orientation}
W.~T. Freeman and M.~Roth, ``Orientation histograms for hand gesture recognition,'' in \emph{International workshop on automatic face and gesture recognition}, vol.~12.\hskip 1em plus 0.5em minus 0.4em\relax Citeseer, 1995, pp. 296--301.

\bibitem{sun2013discriminative}
C.~Sun, T.~Zhang, B.-K. Bao, C.~Xu, and T.~Mei, ``Discriminative exemplar coding for sign language recognition with kinect,'' \emph{IEEE Transactions on Cybernetics}, vol.~43, no.~5, pp. 1418--1428, 2013.

\bibitem{tunga2021pose}
A.~Tunga, S.~V. Nuthalapati, and J.~Wachs, ``Pose-based sign language recognition using gcn and bert,'' in \emph{Proceedings of the IEEE/CVF winter conference on applications of computer vision}, 2021, pp. 31--40.

\bibitem{hu2021signbert}
H.~Hu, W.~Zhao, W.~Zhou, Y.~Wang, and H.~Li, ``Signbert: pre-training of hand-model-aware representation for sign language recognition,'' in \emph{Proceedings of the IEEE/CVF international conference on computer vision}, 2021, pp. 11\,087--11\,096.

\bibitem{vac}
Y.~Min, A.~Hao, X.~Chai, and X.~Chen, ``Visual alignment constraint for continuous sign language recognition,'' in \emph{2021 IEEE/CVF International Conference on Computer Vision (ICCV)}, 2021, pp. 11\,522--11\,531.

\bibitem{smkd}
\BIBentryALTinterwordspacing
A.~Hao, Y.~Min, and X.~Chen, ``Self-mutual distillation learning for continuous sign language recognition,'' \emph{2021 IEEE/CVF International Conference on Computer Vision (ICCV)}, pp. 11\,283--11\,292, 2021. [Online]. Available: \url{https://api.semanticscholar.org/CorpusID:244073884}
\BIBentrySTDinterwordspacing

\bibitem{corrnet}
L.~Hu, L.~Gao, Z.~Liu, and W.~Feng, ``Continuous sign language recognition with correlation network,'' in \emph{2023 IEEE/CVF Conference on Computer Vision and Pattern Recognition (CVPR)}, 2023, pp. 2529--2539.

\bibitem{twostream}
\BIBentryALTinterwordspacing
Y.~Chen, R.~Zuo, F.~Wei, Y.~Wu, S.~Liu, and B.~K.-W. Mak, ``Two-stream network for sign language recognition and translation,'' \emph{ArXiv}, vol. abs/2211.01367, 2022. [Online]. Available: \url{https://api.semanticscholar.org/CorpusID:253254833}
\BIBentrySTDinterwordspacing

\bibitem{tcnet}
\BIBentryALTinterwordspacing
H.~Lu, A.~A. Salah, and R.~Poppe, ``Tcnet: Continuous sign language recognition from trajectories and correlated regions,'' 2024. [Online]. Available: \url{https://api.semanticscholar.org/CorpusID:268531355}
\BIBentrySTDinterwordspacing

\bibitem{corrnet+}
L.~Hu, W.~Feng, L.~Gao, Z.~Liu, and L.~Wan, ``Corrnet+: Sign language recognition and translation via spatial-temporal correlation,'' \emph{arXiv preprint arXiv:2404.11111}, 2024.

\bibitem{jia2016dynamic}
X.~Jia, B.~De~Brabandere, T.~Tuytelaars, and L.~V. Gool, ``Dynamic filter networks,'' \emph{Advances in neural information processing systems}, vol.~29, 2016.

\bibitem{chen2020dynamic}
Y.~Chen, X.~Dai, M.~Liu, D.~Chen, L.~Yuan, and Z.~Liu, ``Dynamic convolution: Attention over convolution kernels,'' in \emph{Proceedings of the IEEE/CVF conference on computer vision and pattern recognition}, 2020, pp. 11\,030--11\,039.

\bibitem{odconv}
C.~Li, A.~Zhou, and A.~Yao, ``Omni-dimensional dynamic convolution,'' \emph{arXiv preprint arXiv:2209.07947}, 2022.

\bibitem{tada1}
Z.~Huang, S.~Zhang, L.~Pan, Z.~Qing, M.~Tang, Z.~Liu, and M.~H. Ang~Jr, ``Tada! temporally-adaptive convolutions for video understanding,'' \emph{arXiv preprint arXiv:2110.06178}, 2021.

\bibitem{tada2}
Z.~Huang, S.~Zhang, L.~Pan, Z.~Qing, Y.~Zhang, Z.~Liu, and M.~H. Ang~Jr, ``Temporally-adaptive models for efficient video understanding,'' \emph{arXiv preprint arXiv:2308.05787}, 2023.

\bibitem{cvsign}
Y.~Yu, S.~Liu, Y.~Feng, M.~Xu, Z.~Jin, and X.~Yang, ``Improving continuous sign language recognition via cross-frame interactions in expanded contextual spaces,'' in \emph{ICASSP 2025-2025 IEEE International Conference on Acoustics, Speech and Signal Processing (ICASSP)}.\hskip 1em plus 0.5em minus 0.4em\relax IEEE, 2025, pp. 1--5.

\bibitem{olmd}
------, ``Olmd: Orientation-aware long-term motion decoupling for continuous sign language recognition,'' in \emph{Proceedings of the AAAI Conference on Artificial Intelligence}, vol.~39, no.~9, 2025, pp. 9707--9715.

\bibitem{tam}
Z.~Liu, L.~Wang, W.~Wu, C.~Qian, and T.~Lu, ``Tam: Temporal adaptive module for video recognition,'' in \emph{Proceedings of the IEEE/CVF international conference on computer vision}, 2021, pp. 13\,708--13\,718.

\bibitem{sen}
\BIBentryALTinterwordspacing
L.~Hu, L.~Gao, Z.~Liu, and W.~Feng, ``Self-emphasizing network for continuous sign language recognition,'' in \emph{AAAI Conference on Artificial Intelligence}, 2022. [Online]. Available: \url{https://api.semanticscholar.org/CorpusID:254096222}
\BIBentrySTDinterwordspacing

\bibitem{dnf}
\BIBentryALTinterwordspacing
R.~Cui, H.~Liu, and C.~Zhang, ``A deep neural framework for continuous sign language recognition by iterative training,'' \emph{IEEE Transactions on Multimedia}, vol.~21, pp. 1880--1891, 2019. [Online]. Available: \url{https://api.semanticscholar.org/CorpusID:68149654}
\BIBentrySTDinterwordspacing

\bibitem{Pu2019IterativeAN}
\BIBentryALTinterwordspacing
J.~Pu, W.~gang Zhou, and H.~Li, ``Iterative alignment network for continuous sign language recognition,'' \emph{2019 IEEE/CVF Conference on Computer Vision and Pattern Recognition (CVPR)}, pp. 4160--4169, 2019. [Online]. Available: \url{https://api.semanticscholar.org/CorpusID:195443370}
\BIBentrySTDinterwordspacing

\bibitem{radialctc}
Y.~Min, P.~Jiao, Y.~Li, W.~Xiaotao, L.~LEI, X.~Chai, and X.~Chen, ``Deep radial embedding for visual sequence learning.'' vol. 13666, 2022, pp. 240--256.

\bibitem{tlp}
\BIBentryALTinterwordspacing
L.~Hu, L.~Gao, Z.~Liu, and W.~Feng, ``Temporal lift pooling for continuous sign language recognition,'' in \emph{European Conference on Computer Vision}, 2022. [Online]. Available: \url{https://api.semanticscholar.org/CorpusID:250626845}
\BIBentrySTDinterwordspacing

\bibitem{lee2021intermediate}
J.~Lee and S.~Watanabe, ``Intermediate loss regularization for ctc-based speech recognition,'' in \emph{ICASSP 2021-2021 IEEE International Conference on Acoustics, Speech and Signal Processing (ICASSP)}.\hskip 1em plus 0.5em minus 0.4em\relax IEEE, 2021, pp. 6224--6228.

\bibitem{cosign}
P.~Jiao, Y.~Min, Y.~Li, X.~Wang, L.~Lei, and X.~Chen, ``Cosign: Exploring co-occurrence signals in skeleton-based continuous sign language recognition,'' in \emph{2023 IEEE/CVF International Conference on Computer Vision (ICCV)}, 2023, pp. 20\,619--20\,629.

\bibitem{cvtslr}
J.~Zheng, Y.~Wang, C.~Tan, S.~Li, G.~Wang, J.~Xia, Y.~Chen, and S.~Z. Li, ``Cvt-slr: Contrastive visual-textual transformation for sign language recognition with variational alignment,'' in \emph{2023 IEEE/CVF Conference on Computer Vision and Pattern Recognition (CVPR)}, 2023, pp. 23\,141--23\,150.

\bibitem{stmc}
H.~Zhou, W.~Zhou, Y.~Zhou, and H.~Li, ``Spatial-temporal multi-cue network for sign language recognition and translation,'' \emph{IEEE Transactions on Multimedia}, vol.~24, pp. 768--779, 2022.

\bibitem{hog}
P.~Buehler, A.~Zisserman, and M.~Everingham, ``Learning sign language by watching tv (using weakly aligned subtitles),'' in \emph{2009 IEEE Conference on Computer Vision and Pattern Recognition}.\hskip 1em plus 0.5em minus 0.4em\relax IEEE, 2009, pp. 2961--2968.

\bibitem{fourier}
\BIBentryALTinterwordspacing
M.~AL-Rousan, K.~Assaleh, and A.~Tala’a, ``Video-based signer-independent arabic sign language recognition using hidden markov models,'' \emph{Applied Soft Computing}, vol.~9, no.~3, pp. 990--999, 2009. [Online]. Available: \url{https://www.sciencedirect.com/science/article/pii/S1568494609000209}
\BIBentrySTDinterwordspacing

\bibitem{c2slr}
R.~Zuo and B.~Mak, ``C2slr: Consistency-enhanced continuous sign language recognition,'' in \emph{2022 IEEE/CVF Conference on Computer Vision and Pattern Recognition (CVPR)}, 2022, pp. 5121--5130.

\bibitem{c2st}
H.~Zhang, Z.~Guo, Y.~Yang, X.~Liu, and D.~Hu, ``C2st: Cross-modal contextualized sequence transduction for continuous sign language recognition,'' in \emph{Proceedings of the IEEE/CVF International Conference on Computer Vision}, 2023, pp. 21\,053--21\,062.

\bibitem{jointctc}
S.~Kim, T.~Hori, and S.~Watanabe, ``Joint ctc-attention based end-to-end speech recognition using multi-task learning,'' in \emph{2017 IEEE international conference on acoustics, speech and signal processing (ICASSP)}.\hskip 1em plus 0.5em minus 0.4em\relax IEEE, 2017, pp. 4835--4839.

\bibitem{autoasr}
P.~Ma, A.~Haliassos, A.~Fernandez-Lopez, H.~Chen, S.~Petridis, and M.~Pantic, ``Auto-avsr: Audio-visual speech recognition with automatic labels,'' in \emph{ICASSP 2023-2023 IEEE International Conference on Acoustics, Speech and Signal Processing (ICASSP)}.\hskip 1em plus 0.5em minus 0.4em\relax IEEE, 2023, pp. 1--5.

\bibitem{ma2021end}
P.~Ma, S.~Petridis, and M.~Pantic, ``End-to-end audio-visual speech recognition with conformers,'' in \emph{ICASSP 2021-2021 IEEE International Conference on Acoustics, Speech and Signal Processing (ICASSP)}.\hskip 1em plus 0.5em minus 0.4em\relax IEEE, 2021, pp. 7613--7617.

\bibitem{nozaki2021relaxing}
J.~Nozaki and T.~Komatsu, ``Relaxing the conditional independence assumption of ctc-based asr by conditioning on intermediate predictions,'' \emph{arXiv preprint arXiv:2104.02724}, 2021.

\bibitem{koller2017re}
O.~Koller, S.~Zargaran, and H.~Ney, ``Re-sign: Re-aligned end-to-end sequence modelling with deep recurrent cnn-hmms,'' in \emph{Proceedings of the IEEE conference on computer vision and pattern recognition}, 2017, pp. 4297--4305.

\bibitem{koller2019weakly}
O.~Koller, N.~C. Camgoz, H.~Ney, and R.~Bowden, ``Weakly supervised learning with multi-stream cnn-lstm-hmms to discover sequential parallelism in sign language videos,'' \emph{IEEE transactions on pattern analysis and machine intelligence}, vol.~42, no.~9, pp. 2306--2320, 2019.

\bibitem{ctca}
\BIBentryALTinterwordspacing
L.~Guo, W.~Xue, Q.~Guo, B.~Liu, K.~Zhang, T.~Yuan, and S.~Chen, ``Distilling cross-temporal contexts for continuous sign language recognition,'' \emph{2023 IEEE/CVF Conference on Computer Vision and Pattern Recognition (CVPR)}, pp. 10\,771--10\,780, 2023. [Online]. Available: \url{https://api.semanticscholar.org/CorpusID:261081454}
\BIBentrySTDinterwordspacing

\bibitem{signvtcl}
H.~Chen, J.~Wang, Z.~Guo, J.~Li, D.~Zhou, B.~Wu, C.~Guan, G.~Chen, and P.-A. Heng, ``Signvtcl: multi-modal continuous sign language recognition enhanced by visual-textual contrastive learning,'' \emph{arXiv preprint arXiv:2401.11847}, 2024.

\bibitem{yang2019condconv}
B.~Yang, G.~Bender, Q.~V. Le, and J.~Ngiam, ``Condconv: Conditionally parameterized convolutions for efficient inference,'' \emph{Advances in neural information processing systems}, vol.~32, 2019.

\bibitem{bivolution}
X.~Hu, X.~Chen, B.~Ni, T.~Li, and Y.~Liu, ``Bi-volution: a static and dynamic coupled filter,'' in \emph{Proceedings of the AAAI Conference on Artificial Intelligence}, vol.~36, no.~1, 2022, pp. 960--968.

\bibitem{klein2015dynamic}
B.~Klein, L.~Wolf, and Y.~Afek, ``A dynamic convolutional layer for short range weather prediction,'' in \emph{Proceedings of the IEEE Conference on Computer Vision and Pattern Recognition}, 2015, pp. 4840--4848.

\bibitem{ha2016hypernetworks}
D.~Ha, A.~Dai, and Q.~V. Le, ``Hypernetworks,'' \emph{arXiv preprint arXiv:1609.09106}, 2016.

\bibitem{ctc}
A.~Graves, S.~Fern{\'a}ndez, F.~Gomez, and J.~Schmidhuber, ``Connectionist temporal classification: labelling unsegmented sequence data with recurrent neural networks,'' in \emph{Proceedings of the 23rd international conference on Machine learning}, 2006, pp. 369--376.

\bibitem{nakagome2022interaug}
Y.~Nakagome, T.~Komatsu, Y.~Fujita, S.~Ichimura, and Y.~Kida, ``Interaug: augmenting noisy intermediate predictions for ctc-based asr,'' \emph{arXiv preprint arXiv:2204.00174}, 2022.

\bibitem{crctc}
Z.~Yao, W.~Kang, X.~Yang, F.~Kuang, L.~Guo, H.~Zhu, Z.~Jin, Z.~Li, L.~Lin, and D.~Povey, ``Cr-ctc: Consistency regularization on ctc for improved speech recognition,'' \emph{arXiv preprint arXiv:2410.05101}, 2024.

\bibitem{yu2021boundary}
F.~Yu, H.~Luo, P.~Guo, Y.~Liang, Z.~Yao, L.~Xie, Y.~Gao, L.~Hou, and S.~Zhang, ``Boundary and context aware training for cif-based non-autoregressive end-to-end asr,'' in \emph{2021 IEEE Automatic Speech Recognition and Understanding Workshop (ASRU)}.\hskip 1em plus 0.5em minus 0.4em\relax IEEE, 2021, pp. 328--334.

\bibitem{fan2021cass}
R.~Fan, W.~Chu, P.~Chang, and J.~Xiao, ``Cass-nat: Ctc alignment-based single step non-autoregressive transformer for speech recognition,'' in \emph{ICASSP 2021-2021 IEEE International Conference on Acoustics, Speech and Signal Processing (ICASSP)}.\hskip 1em plus 0.5em minus 0.4em\relax IEEE, 2021, pp. 5889--5893.

\bibitem{han2023knowledge}
M.~Han, F.~Chen, J.~Shi, S.~Xu, and B.~Xu, ``Knowledge transfer from pre-trained language models to cif-based speech recognizers via hierarchical distillation,'' \emph{arXiv preprint arXiv:2301.13003}, 2023.

\bibitem{dong2020cif}
L.~Dong and B.~Xu, ``Cif: Continuous integrate-and-fire for end-to-end speech recognition,'' in \emph{ICASSP 2020-2020 IEEE International Conference on Acoustics, Speech and Signal Processing (ICASSP)}.\hskip 1em plus 0.5em minus 0.4em\relax IEEE, 2020, pp. 6079--6083.

\bibitem{hori2017advances}
T.~Hori, S.~Watanabe, Y.~Zhang, and W.~Chan, ``Advances in joint ctc-attention based end-to-end speech recognition with a deep cnn encoder and rnn-lm,'' \emph{arXiv preprint arXiv:1706.02737}, 2017.

\bibitem{senet}
J.~Hu, L.~Shen, and G.~Sun, ``Squeeze-and-excitation networks,'' in \emph{Proceedings of the IEEE conference on computer vision and pattern recognition}, 2018, pp. 7132--7141.

\bibitem{subunet}
\BIBentryALTinterwordspacing
N.~C. Camg{\"o}z, S.~Hadfield, O.~Koller, and R.~Bowden, ``Subunets: End-to-end hand shape and continuous sign language recognition,'' \emph{2017 IEEE International Conference on Computer Vision (ICCV)}, pp. 3075--3084, 2017. [Online]. Available: \url{https://api.semanticscholar.org/CorpusID:38408000}
\BIBentrySTDinterwordspacing

\bibitem{Hu2023AdaBrowseAV}
\BIBentryALTinterwordspacing
L.~Hu, L.~Gao, Z.~Liu, C.-M. Pun, and W.~Feng, ``Adabrowse: Adaptive video browser for efficient continuous sign language recognition,'' \emph{Proceedings of the 31st ACM International Conference on Multimedia}, 2023. [Online]. Available: \url{https://api.semanticscholar.org/CorpusID:260926078}
\BIBentrySTDinterwordspacing

\bibitem{slowfastsign}
J.~Ahn, Y.~Jang, and J.~S. Chung, ``Slowfast network for continuous sign language recognition,'' in \emph{IEEE International Conference on Acoustics, Speech and Signal Processing}, 2024, pp. 3920--3924.

\bibitem{signgraph}
S.~Gan, Y.~Yin, Z.~Jiang, H.~Wen, L.~Xie, and S.~Lu, ``Signgraph: A sign sequence is worth graphs of nodes,'' in \emph{Proceedings of the IEEE/CVF Conference on Computer Vision and Pattern Recognition (CVPR)}.\hskip 1em plus 0.5em minus 0.4em\relax IEEE, 2024.

\bibitem{signbert}
H.~Hu, W.~Zhao, W.~Zhou, and H.~Li, ``Signbert+: Hand-model-aware self-supervised pre-training for sign language understanding,'' \emph{IEEE Transactions on Pattern Analysis \&; Machine Intelligence}, vol.~45, no.~09, pp. 11\,221--11\,239, sep 2023.

\bibitem{GPDN}
L.~Guo, W.~Xue, B.~Liu, K.~Zhang, T.~Yuan, and D.~Metaxas, ``Gloss prior guided visual feature learning for continuous sign language recognition,'' \emph{IEEE Transactions on Image Processing}, vol.~33, pp. 3486--3495, 2024.

\bibitem{Forster2015Continuous}
Forster, Jens, Ney, Hermann, Koller, and Oscar, ``Continuous sign language recognition: Towards large vocabulary statistical recognition systems handling multiple signers,'' \emph{Computer vision and image understanding: CVIU}, vol. 141, pp. 108--125, 2015.

\bibitem{8578910}
N.~C. Camgoz, S.~Hadfield, O.~Koller, H.~Ney, and R.~Bowden, ``Neural sign language translation,'' in \emph{2018 IEEE/CVF Conference on Computer Vision and Pattern Recognition}, 2018, pp. 7784--7793.

\bibitem{9578398}
H.~Zhou, W.~Zhou, W.~Qi, J.~Pu, and H.~Li, ``Improving sign language translation with monolingual data by sign back-translation,'' in \emph{2021 IEEE/CVF Conference on Computer Vision and Pattern Recognition (CVPR)}, 2021, pp. 1316--1325.

\bibitem{resnet}
\BIBentryALTinterwordspacing
K.~He, X.~Zhang, S.~Ren, and J.~Sun, ``Deep residual learning for image recognition,'' \emph{2016 IEEE Conference on Computer Vision and Pattern Recognition (CVPR)}, pp. 770--778, 2015. [Online]. Available: \url{https://api.semanticscholar.org/CorpusID:206594692}
\BIBentrySTDinterwordspacing

\bibitem{imagenet}
\BIBentryALTinterwordspacing
J.~Deng, W.~Dong, R.~Socher, L.-J. Li, K.~Li, and L.~Fei-Fei, ``Imagenet: A large-scale hierarchical image database,'' \emph{2009 IEEE Conference on Computer Vision and Pattern Recognition}, pp. 248--255, 2009. [Online]. Available: \url{https://api.semanticscholar.org/CorpusID:57246310}
\BIBentrySTDinterwordspacing

\bibitem{lshan}
\BIBentryALTinterwordspacing
J.~Huang, W.~gang Zhou, Q.~Zhang, H.~Li, and W.~Li, ``Video-based sign language recognition without temporal segmentation,'' in \emph{AAAI Conference on Artificial Intelligence}, 2018. [Online]. Available: \url{https://api.semanticscholar.org/CorpusID:9005234}
\BIBentrySTDinterwordspacing

\bibitem{alyami2025swin}
S.~Alyami and H.~Luqman, ``Swin-mstp: Swin transformer with multi-scale temporal perception for continuous sign language recognition,'' \emph{Neurocomputing}, vol. 617, p. 129015, 2025.

\bibitem{unisign}
Z.~Li, W.~Zhou, W.~Zhao, K.~Wu, H.~Hu, and H.~Li, ``Uni-sign: Toward unified sign language understanding at scale,'' \emph{arXiv preprint arXiv:2501.15187}, 2025.

\bibitem{slowfast}
C.~Feichtenhofer, H.~Fan, J.~Malik, and K.~He, ``Slowfast networks for video recognition,'' in \emph{Proceedings of the IEEE/CVF international conference on computer vision}, 2019, pp. 6202--6211.

\bibitem{xie2018rethinking}
S.~Xie, C.~Sun, J.~Huang, Z.~Tu, and K.~Murphy, ``Rethinking spatiotemporal feature learning: Speed-accuracy trade-offs in video classification,'' in \emph{Proceedings of the European conference on computer vision (ECCV)}, 2018, pp. 305--321.

\bibitem{gradcam}
\BIBentryALTinterwordspacing
R.~R. Selvaraju, A.~Das, R.~Vedantam, M.~Cogswell, D.~Parikh, and D.~Batra, ``Grad-cam: Visual explanations from deep networks via gradient-based localization,'' \emph{International Journal of Computer Vision}, vol. 128, pp. 336 -- 359, 2016. [Online]. Available: \url{https://api.semanticscholar.org/CorpusID:15019293}
\BIBentrySTDinterwordspacing

\bibitem{dse}
S.~Wang, L.~Guo, and W.~Xue, ``Dynamical semantic enhancement network for continuous sign language recognition,'' \emph{Multimedia Systems}, vol.~30, p. 313, 2024.

\bibitem{squeezenet}
F.~N. Iandola, S.~Han, M.~W. Moskewicz, K.~Ashraf, W.~J. Dally, and K.~Keutzer, ``Squeezenet: Alexnet-level accuracy with 50x fewer parameters and< 0.5 mb model size,'' \emph{arXiv preprint arXiv:1602.07360}, 2016.

\bibitem{dla}
F.~Yu, D.~Wang, E.~Shelhamer, and T.~Darrell, ``Deep layer aggregation,'' in \emph{Proceedings of the IEEE conference on computer vision and pattern recognition}, 2018, pp. 2403--2412.

\end{thebibliography}
}
\vspace{-\baselineskip}
\clearpage

\twocolumn[
  \begin{center}
    \LARGE \bfseries Supplementary Materials
    \vspace{1.0em}
  \end{center}
]


This supplementary material includes additional details not shown in the main paper. Specifically, Sec. \ref{sec:1} presents a detailed analysis of the computational complexity and parameter count of DCAC, along with the actual complexity and parameterization of DESign. Sec. \ref{sec:2} derives and analyzes the underlying causes of CTC spikiness, while Sec. \ref{sec:3} demonstrates the effectiveness of SR-CTC based on these findings. Sec. \ref{sec:4}, \ref{sec:5}, and \ref{sec:6} report additional comparative experiments, ablations, and visualizations. Finally, Sec. \ref{sec:7} discusses the limitations of our work and outlines potential directions for future research.

\section{Derivation: Time Complexity and Parameters}
\label{sec:1}
\subsection{An Overview of FLOPs and Parameters in DCAC }

To quantify the computational complexity of the DCAC, we conduct a complexity analysis. For the static branch, it contains only one static convolution, so the complexity and parameters are:
\begin{equation}
\label{eq:7}
\begin{aligned}
 &\mathrm{FLOPs}(\mathrm{Static}) =C_{o}\times \frac{C_{i}}{G}\times k_tk_hk_w\times THW ,\\
 &  \mathrm{Params}(\mathrm{Static})=C_o\times \frac{C_i}{G}\times k_tk_hk_w.
\end{aligned}
\end{equation}
 For the dynamic branches, the computational load is primarily attributed to Conv2 and the 3D convolution operations, while the majority of the parameter count stems from the Conv2. Therefore, we have:
\begin{equation}
\label{eq:8}
\begin{aligned}
 &\mathrm{FLOPs}(\mathrm{Dynamic}) \approx C_iT\left(C_iHW+\frac{C_ok_tk_hk_w}{G}\left(C_i+HW\right)\right),\\
 &  \mathrm{Params}(\mathrm{Dynamic}) \approx C_i^2 \left(1 + \frac{C_o k_h k_w}{G} \right),
\end{aligned}
\end{equation}
The overall computational load and parameter estimates for DCAC are:
\begin{equation}
\label{eq:8}
\begin{aligned}
\mathrm{FLOPs}(\mathrm{DCAC}) 
&\approx C_iT\left(C_iHW+\frac{C_ok_tk_hk_w}{G}\left(C_i+2HW\right)\right), \\
\mathrm{Params}(\mathrm{DCAC}) 
&\approx \frac{C_i C_o k_h k_w}{G} \times (C_i + k_t)+C_i^2.
\end{aligned}
\end{equation}

\subsection{Detailed Analysis of Dynamic Branch}

 Here, we provide a detailed derivation for both the number of FLOPs and trainable parameters in the Dynamic Branch.

\subsubsection{FLOPs}

The following lists all FLOPs involved in the DCAC dynamic branch, corresponding to Alg. 1 and Fig. 3 in the main paper:

\begin{equation}
\label{eq:flops}
\begin{aligned}
&\text{FLOPs}(\text{Unfold}) = C_i \times T \times H \times W \times k_t k_h k_w, \\
&\text{FLOPs}(\text{GAP}) = C_i \times T \times H \times W, \\
&\text{FLOPs}(\text{FC}) = C_i \times \frac{C_i}{r} \times T, \\
&\text{FLOPs}(\text{BN})  = \frac{C_i}{r} \times T, \\
&\text{FLOPs}(\text{FCs}) = \frac{C_i}{r} \times \left( \frac{C_i}{G} + C_o + k_t + n \right) \times T, \\
&\text{FLOPs}(\text{Mul1}) = n \times C_o \times \frac{C_i}{G} \times T \times k_t k_h k_w, \\
&\text{FLOPs}(\text{Conv1}) = C_i^2 \times T \times H \times W, \\
&\text{FLOPs}(\text{GAP}) = C_i \times T \times H \times W, \\
&\text{FLOPs}(\text{Unfold}) = C_i \times T \times k_t, \\
&\text{FLOPs}(\text{Conv2}) = \frac{C_i^2}{G} \times C_o \times T \times k_t k_h k_w, \\
&\text{FLOPs}(\text{Mul2}) = \frac{C_i}{G} \times C_o \times T \times k_t k_h k_w, \\
&\text{FLOPs}(\text{Conv3D}) = \frac{C_i}{G} \times C_o \times T \times H \times W \times k_t k_h k_w.
\end{aligned}
\end{equation}

We now analyze and simplify the dominant FLOPs terms. In practical applications, the condition $C_i = C_o = G \gg \{n, k_t, k_h, k_w\}$ generally holds, allowing us to identify the dominant terms: \textbf{Conv1}, \textbf{Conv2}, and \textbf{Conv3D} are the most computationally expensive components.

\paragraph{Dominant Terms}
\begin{itemize}
  \item \textbf{Conv1:}
  \[
  \mathrm{FLOPs}(\mathrm{Conv1}) = C_i^2 \times T \times H\times W.
  \]
  \item \textbf{Conv2:}
  \[
  \mathrm{FLOPs}(\mathrm{Conv2}) = \frac{C_i^2}{G} \times C_o \times T \times k_t k_h k_w .
  \]
  
  \item \textbf{Conv3D:}
  \[
  \mathrm{FLOPs}(\mathrm{Conv3D}) = \frac{C_i}{G} \times C_o \times T \times H \times W \times k_t k_h k_w .
  \]
\end{itemize}
We factor out the common terms and derive the final approximation:
\begin{equation}
\begin{aligned}
\mathrm{FLOPs}(\text{Dynamic}) &\approx C_i^2 \times T \times H\times W+\frac{C_i^2}{G} \times C_o \times T \times k_t k_h k_w \\
&\quad + \frac{C_i}{G} \times C_o \times T \times H \times W \times k_t k_h k_w \\
&= C_i^2 \times THW+\frac{C_i^2C_o}{G} \times T \times k_tk_hk_w \\&
+\frac{C_iC_o}{G}\times THW\times k_tk_hk_w\\&=C_iT\left(C_iHW+\frac{C_ok_tk_hk_w}{G}\left(C_i+HW\right)\right).
\end{aligned}
\end{equation}
This expression captures the dominant computational cost of the dynamic branch.

\subsubsection{Parameters}

According to Fig. 3 in the main text, the parameter count of each trainable module is as follows:
\begin{equation}
\label{eq:params}
\begin{aligned}
&\text{Params}(\text{FC}) = C_i \times \frac{C_i}{r}, \\
&\text{Params}(\text{BN})  = \frac{C_i}{r}, \\
&\text{Params}(\text{FCs}) = \frac{C_i}{r} \times \left( \frac{C_i}{G} + C_o + k_t + n \right), \\
&\text{Params}(\text{Experts}) = n \times C_o \times \frac{C_i}{G} \times k_t k_h k_w, \\
&\text{Params}(\text{Conv1}) = C_i^2, \\
&\text{Params}(\text{Conv2}) = \frac{C_i^2}{G} \times C_o \times k_h k_w.
\end{aligned}
\end{equation}
Summing all terms, we get:
\begin{equation}
\begin{aligned}
\text{Params}(\text{Dynamic}) =\ &\frac{C_i^2}{r} + \frac{C_i}{r} + \frac{C_i}{r} \left( \frac{C_i}{G} + C_o + k_t + n \right) + \\
&\frac{n C_o C_i k_t k_h k_w}{G} + C_i^2 + \frac{C_i^2 C_o k_h k_w}{G}.
\end{aligned}
\end{equation}
First, we can omit the linear terms $\frac{C_i}{r}$ and $\frac{C_i^2}{r}$ since they are negligible compared to $C_i^2$. 
Moreover, given that $\{k_t, n\} \ll C_o$, the following approximation holds:
\begin{equation}
\begin{aligned}
\text{Params}(\text{Dynamic}) \approx\ 
& \frac{C_i}{r} \left( \frac{C_i}{G} + C_o \right) + 
\frac{n C_o C_i k_t k_h k_w}{G} \\&+ 
C_i^2 + 
\frac{C_i^2 C_o k_h k_w}{G}.
\end{aligned}
\end{equation}

Considering the practical setting where $C_i = C_o = G \gg \{n, k_t, k_h, k_w\}$, we can further simplify:

\begin{equation}
\begin{aligned}
\text{Params}(\text{Dynamic}) \approx\ 
& C_i^2 + \frac{C_i^2 C_o k_h k_w}{G} \\
=&\ C_i^2 \left(1 + \frac{C_o k_h k_w}{G} \right).
\end{aligned}
\end{equation}

\section{Peaky Prediction and Overfitting to Dominant Path in CTC}
\label{sec:2}

The peak of CTC is one of the main reasons limiting current CSLR technology. We provide a supplementary analysis of the underlying causes.

\paragraph{CTC Loss Function} Given the logits of model $Z$ and corresponding target sequence $\mathcal{G}$, the CTC loss is defined as:
\begin{equation}
\mathcal{L}_{\mathrm{CTC}} = -\log \sum_{\pi \in \mathcal{B}^{-1}(\mathcal{G})}p(\pi \mid \mathbf{Z}) = -\log \sum_{\pi \in \mathcal{B}^{-1}(\mathcal{G})} \prod_{t=1}^{T'} p(\pi_t \mid \mathbf{Z}),
\label{eq:ctc_loss}
\end{equation}
where $\mathcal{B}$ is a many-to-one mapping that removes repeated labels and blanks from path $\pi$, the per-frame probabilities $p(\pi_t \mid \mathbf{Z})$  is obtained via a softmax operation over the logits $\mathbf{Z}$.

\paragraph{Gradient of CTC Loss} The error signal with respect to the output probability $p(\pi_t' \mid \mathbf{Z})$ is given by :
\begin{equation}
\frac{\partial \mathcal{L}_{\mathrm{CTC}}}{\partial p(\pi_t' \mid \mathbf{Z})} = - \frac{1}{p(\mathcal{G} \mid \mathbf{Z})} \times \frac{1}{ p(\pi_t' \mid \mathbf{Z})} \sum_{\substack{\pi \in \mathcal{B}^{-1}(l) \\ \pi_t = \pi_t'}} p(\pi \mid \mathbf{Z}).
\label{eq:ctc_grad}
\end{equation}
This gradient is proportional to the total probability of all paths that go through symbol $\pi_t'$ at time $t$.

\paragraph{Mechanism of Overfitting and Peaky Prediction} 
From Eq.~\ref{eq:ctc_grad}, we observe that once a specific path $\pi^\ast$ begins to dominate the overall summation, the probability $p(\pi_t^\ast \mid \mathbf{Z})$ at each timestep $t$ along this path becomes significantly larger than the probabilities $p(\pi_t \mid \mathbf{Z})$ of other feasible paths. As a result, $\pi^\ast$ contributes the most to the gradient of the CTC loss. This observation is also consistent with the visualization in Fig.~4 of the main text. Consequently, the model receives stronger gradient signals that reinforce $\pi^\ast$, further increasing its likelihood in a positive feedback loop. This dynamic causes the model to concentrate its prediction mass on a few specific timesteps and symbols, leading to the characteristic sharp peaks in the output distribution---i.e., the \emph{peaky prediction} phenomenon.

Over time, this can lead to overfitting to a single dominant path, even though many feasible paths exist. The model becomes less sensitive to alternative paths and lacks the exploration necessary to improve alignment robustness.

\begin{figure*}[t!]
  \centering
  \includegraphics[width=\textwidth]{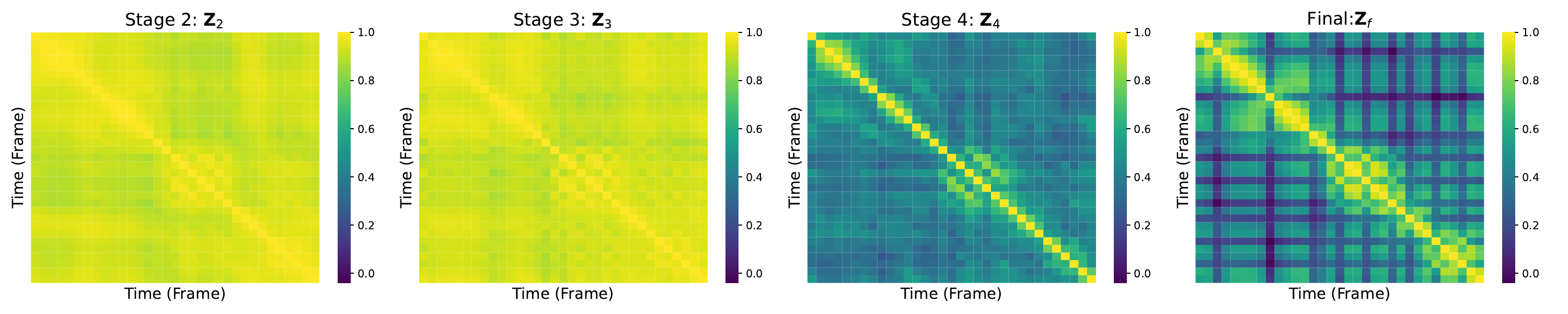}
  \caption{Temporal correlation heatmaps of features from different stages. From left to right, the plots correspond to stage 2, stage 3, stage 4, and the final model output. In earlier stages (e.g., stage 2 and stage 3), the representations exhibit high uniformity across time, reflected by broadly high correlation values and smooth diagonals. This indicates low specialization, where frame-level features are less discriminative. As the model deepens (stage 4 and Final), the correlations become sharper and more structured, with stronger diagonals and distinct temporal patterns.
  }
 \label{fig:heatmap}
\end{figure*}

\section{SR-CTC to Mitigate Dominant Path Overfitting}
\label{sec:3}
Based on the above theory regarding the dominant path phenomenon in CTC, we present a theoretical explanation of how SR-CTC alleviates this issue:

As shown in Fig.~\ref{fig:heatmap}, the representations in earlier stages (e.g., $\mathbf{Z}_2$) are less specialized, and the per-frame probability distributions $p(\pi_t \mid \mathbf{Z}_2)$ tend to be flatter, assigning non-negligible probabilities to a broader range of alignment paths. From the gradient expression in Eq.~\ref{eq:ctc_grad}, we see that the update signal is dominated by paths with large posterior probability. In early layers, since the path distribution is more uniform, the loss gradient is distributed across many paths rather than focused on a single dominant path $\pi^\ast$. This prevents the network from prematurely collapsing its predictions into peaky outputs.

As training progresses, the deeper layers may still develop sharper alignment spikes. However, because the overall training objective includes supervision from earlier layers, the gradient contributions from $\mathcal{L}_{l}$ $(l<L)$ act as a regularization force. These losses from shallower layers encourage the model to maintain a smoother output distribution and explore multiple alignment hypotheses.

 To conclude, supervision from SR-CTC mitigates the tendency of CTC to overfit to a single dominant path. By injecting loss gradients from intermediate representations, the model is encouraged to explore and preserve alternate alignment paths throughout training. This strategy softens the peaky prediction problem and improves temporal alignment robustness.

\section{Additional Comparisions}
\label{sec:4}
\begin{table}[t!]
 \caption{Comparison of existing dynamic convolution methods on the PHOENIX14.}
\center
\resizebox{0.35\textwidth}{!}{
    \begin{tabular}{lcc}
    \toprule
    Methods & Dev(\%) & Test(\%) \\ 
    \midrule 
    -& 19.2& 19.5\\
    w/ CondConv~\cite{yang2019condconv}&22.0 &22.6\\ 
    w/ DyConv~\cite{chen2020dynamic}& 19.0&19.9\\ 
    w/ ODConv~\cite{odconv}& 21.7&21.9\\  
    w/ TAda V1~\cite{tada1} &22.2 &22.5\\
    w/ TAda V2~\cite{tada2} & 19.7& 20.3\\ 
    w/ GIConv2d~\cite{dse} &18.7&19.8\\
    \midrule 
    w/ DCAC (ours)&\textbf{17.9}&\textbf{18.0}\\
    \bottomrule
    \end{tabular}
 }

 \label{tab:dynamics}

\end{table}

\begin{table}[t!]
 \caption{Comparison of existing methods for mitigating CTC peak issues on the PHOENIX14.}
\center
\resizebox{0.35\textwidth}{!}{
    \begin{tabular}{lcc}
    \toprule
    Methods & Dev(\%) & Test(\%) \\ 
    \midrule 
    -& 21.5 & 22.2\\
    w/ VAC~\cite{vac}&19.7 & 20.0 \\ 
    w/ SMKD~\cite{smkd}& 19.7& 20.1\\ 
    \midrule 
    w/ SR-CTC (ours)&\textbf{18.6}&\textbf{19.6}\\
    \bottomrule
    \end{tabular}
 }
 \label{tab:ctcs}
\end{table}

\textbf{Comparison of DCAC with related methods on PHOENIX14.}
As shown in Tab. \ref{tab:dynamics}, we compare DCAC with existing dynamic convolution methods on the PHOENIX14 dataset. DCAC achieves the best performance among all methods, outperforming the previous best approach, GIConv2d \cite{dse}, with significant WER reductions of 0.8\% on the development set and 1.8\% on the test set. This improvement can be attributed to DCAC’s well-designed architecture and its strong capability to adapt convolutional kernels to frame-wise contextual information. These results also highlight the potential of dynamic convolutions in CSLR applications.

\textbf{Comparison of SR-CTC with related methods on PHOENIX14.} 
We compare SR-CTC with two existing methods for mitigating the CTC spike phenomenon in Tab. \ref{tab:ctcs}. To independently compare the effectiveness of VAC \cite{vac}, SMKD \cite{smkd}, and our SR-CTC, we re-established a clean baseline composed of ResNet-34, 1D CNN, and BiLSTM. The results confirm the effectiveness of both VAC and SMKD: VAC reduces the WER by 1.8\% on the Dev set and 2.2\% on the Test set, while SMKD achieves reductions of 1.8\% and 2.1\%, respectively. In contrast, SR-CTC demonstrates even stronger performance, achieving WER reductions of 2.9\% on the Dev set and 2.6\% on the Test set compared to the baseline. Notably, SR-CTC is orthogonal to both VAC and SMKD, and can be integrated alongside them to further enhance the baseline network’s performance without incurring any additional inference overhead.

\textbf{Comparison of parameters and complexity with the baseline and existing methods.} As shown in Table \ref{tab:efficiency}, we report the actual number of parameters, computational complexity, and inference time for all models, including our proposed DESign, the baseline, and previous methods. All tests were conducted on a 100-frame video sample. Compared to prior approaches, our baseline incurs higher computational overhead due to the use of a more powerful ResNet34 as the feature extractor, whereas previous methods typically employ ResNet18. Notably, DESign introduces only a modest increase over the baseline—1.16\% in the number of parameters and 1.15\% in computational complexity. Although the inference latency increases by 9.45 ms, DESign still maintains real-time performance.

\begin{table}[!t]
\centering
\caption{Comparison of computational cost, model size, and inference time (evaluated on a 100-frame video).}
\resizebox{0.5\textwidth}{!}{
\begin{tabular}{lccc}
\toprule
\textbf{Method} & \textbf{FLOPs (G)} & \textbf{Params (M)} & \textbf{Time (ms)} \\
\midrule
CorrNet \cite{corrnet}      & 187.71 & 32.05 & 57.67 \\
CorrNet+ \cite{corrnet+}    & 188.60 & 57.29 & 45.79 \\
SEN  \cite{sen}        & 185.60 & 34.51 & 46.10 \\
TLP  \cite{tlp}        & 184.05 & 59.50  & 31.55 \\
\midrule
Baseline     & 369.46 & 66.95 & 59.98 \\
\textbf{DESign} & 373.75 (+4.29, $\uparrow$ 1.16\%) & 67.72  (+0.77, $\uparrow$ 1.15\%) & 69.43 (+9.45) \\
\bottomrule
\end{tabular}
}
\label{tab:efficiency}
\end{table}

\section{Additional Ablations}
\label{sec:5}
    


\begin{table}[t]
\centering
\caption{Evaluating the impact of DESign on different feature extractors}
\resizebox{0.33\textwidth}{!}{
\begin{tabular}{lcc}
\toprule
\textbf{Configuration} & \textbf{Dev (\%)} & \textbf{Test (\%)} \\
\midrule
SqueezeNet \cite{squeezenet}        & 24.9  & 24.4\\
\quad w/ DESign     & 23.0 ($\downarrow$ 1.9)  & 23.4 ($\downarrow$ 1.0)  \\
\midrule
DLA34 \cite{dla}     &18.9   &19.3 \\
\quad w/ DESign     & 17.5 ($\downarrow$ 1.4) & 17.7 ($\downarrow$ 1.6)\\
\midrule
ResNet18 \cite{resnet}    & 19.6 & 20.2  \\
\quad w/ DESign  &  17.5 ($\downarrow$ 2.1)  & 17.7 ($\downarrow$ 2.5)  \\
\midrule
ResNet34 \cite{resnet}  & 19.2 & 19.5  \\
\quad w/ DESign     & 17.1 ($\downarrow$ 2.1)  & 17.4 ($\downarrow$ 2.1) \\
\bottomrule
\label{tab:fe}
\end{tabular}
}
\end{table}

\textbf{Evaluating the impact of DESign on different feature extractors.}
We evaluate the effectiveness of DESign across different feature extractors and observe consistent reductions in WER across all baselines. Specifically, DESign reduces WER by 1.9\% (Dev) and 1.0\% (Test) on SqueezeNet \cite{squeezenet}, 1.4\% (Dev) and 1.6\% (Test) on DLA34 \cite{dla}, and 2.1\% (Dev) and 2.5\% (Test) on ResNet18 \cite{resnet}. On ResNet34 \cite{resnet}, DESign achieves a consistent 2.1\% reduction on both Dev and Test sets—the best overall performance. As a result, we adopt ResNet34 as our default feature extractor.

\section{More Visualizations}
\label{sec:6}

\begin{figure}[t!]
  \centering
  \includegraphics[width=0.85\linewidth]{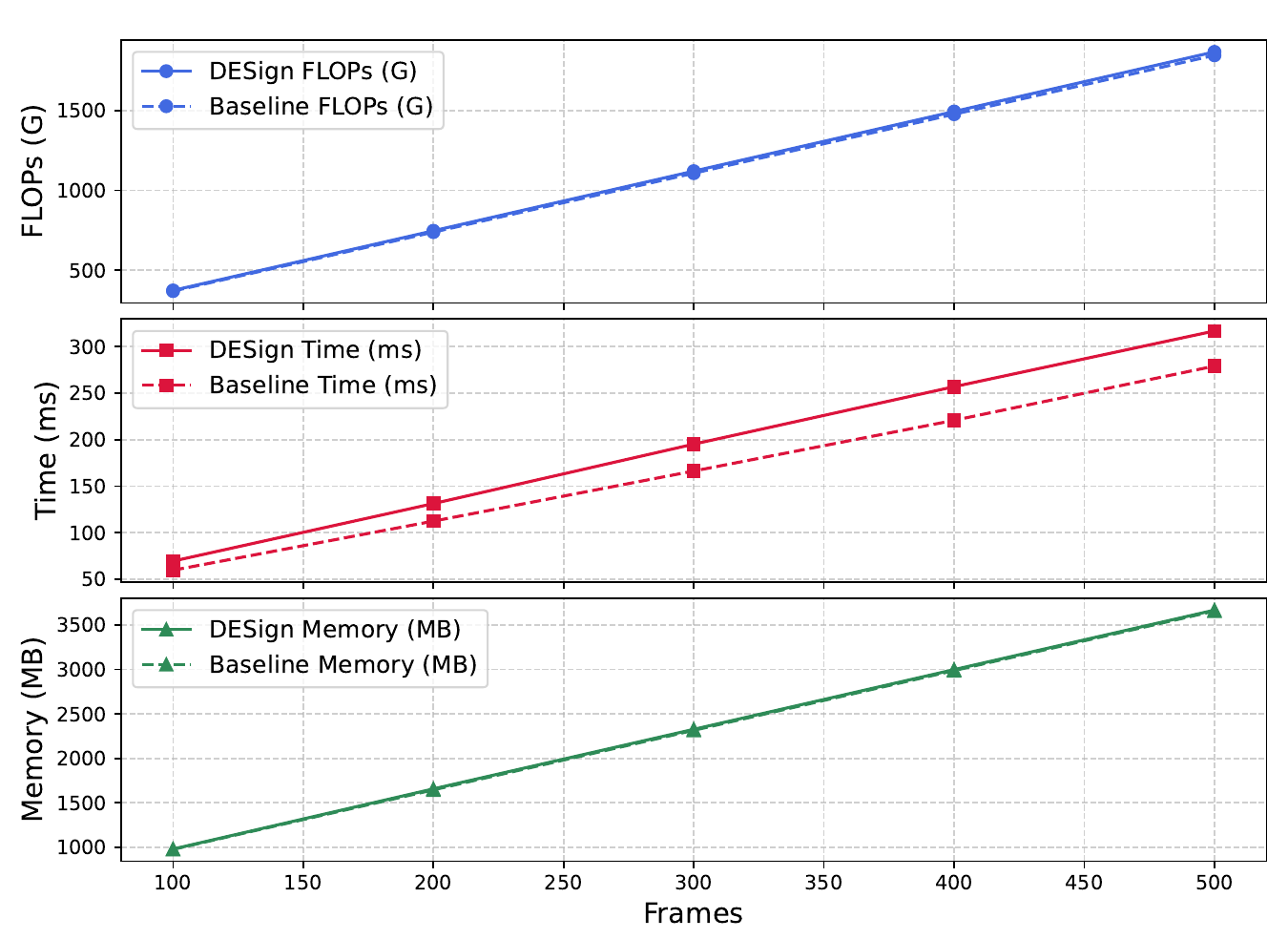}
  \caption{Comparison of FLOPs, inference time, and memory consumption between DESign and Baseline across varying input frame lengths.}
 \label{fig:flops}
\end{figure}


\textbf{Impact of input length on DESign and Baseline.} The frame length of continuous sign language videos is not fixed—typically ranging from 100 to 300 frames in datasets, and potentially longer in real-world applications. Therefore, it is necessary to investigate how DESign's FLOPs, inference time, and memory usage vary with different input frame lengths, and compare these metrics against the Baseline. As shown in Fig. \ref{fig:flops}, all three metrics—FLOPs, inference time, and memory consumption—increase naturally as the number of input frames grows. Notably, the FLOPs and memory curves of DESign are nearly identical to those of the Baseline, indicating that changes in input length affect DESign similarly to the Baseline, and the introduction of the DCAC module does not lead to significant additional resource overhead. Although DESign exhibits slightly higher inference time than the Baseline, the overall increase remains modest, demonstrating its good scalability and practical applicability for deployment.

\begin{figure}[t!]
  \centering
  \includegraphics[width=\linewidth]{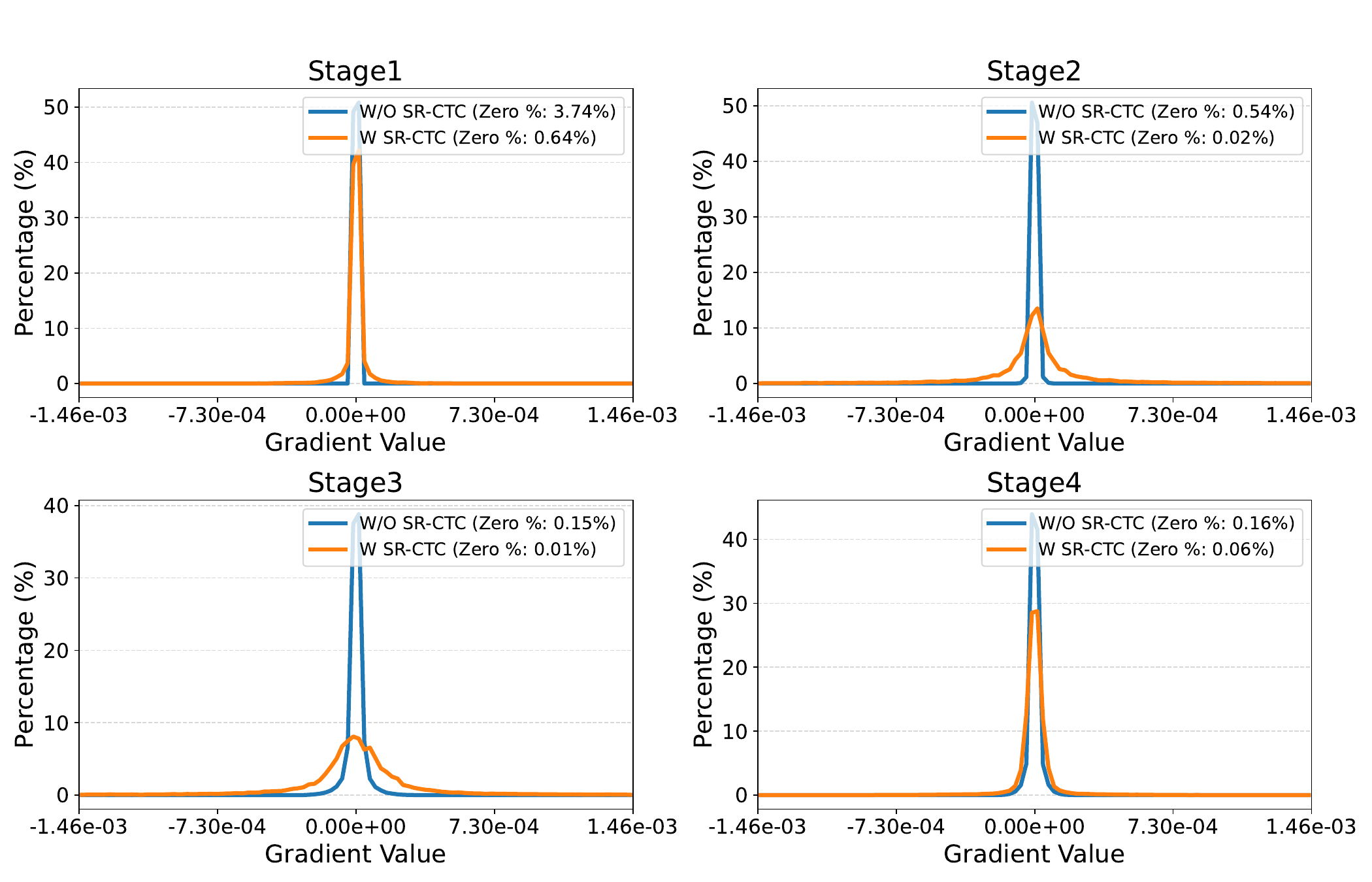}
  \caption{Gradient distribution and zero-value ratio across different stages in the late training phase, comparing models with and without SR-CTC.}
 \label{fig:gdistribution}
\end{figure}

\begin{figure*}[t!]
  \centering
  \includegraphics[width=0.9\textwidth]{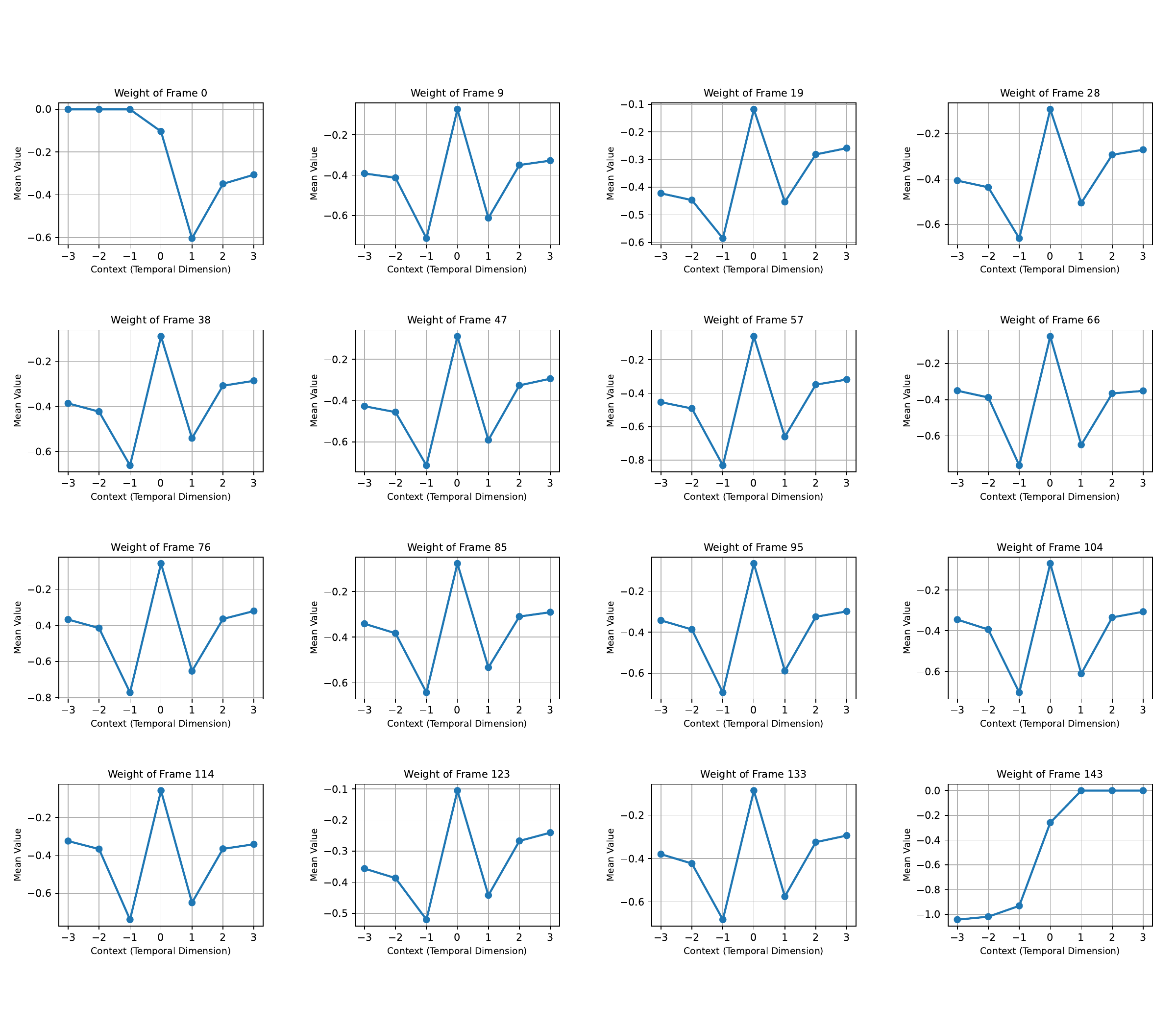}
  \caption{ Visualization of weight variations along the context dimension (set to 7 in this case) generated by CAKG across different frames during inference. 16 frames are uniformly sampled from a 143-frame video. For visualization purposes, average pooling is applied over the channel dimension.}
 \label{fig:cakgv1}
\end{figure*}

\begin{figure}[t!]
  \centering
  \includegraphics[width=0.8\linewidth]{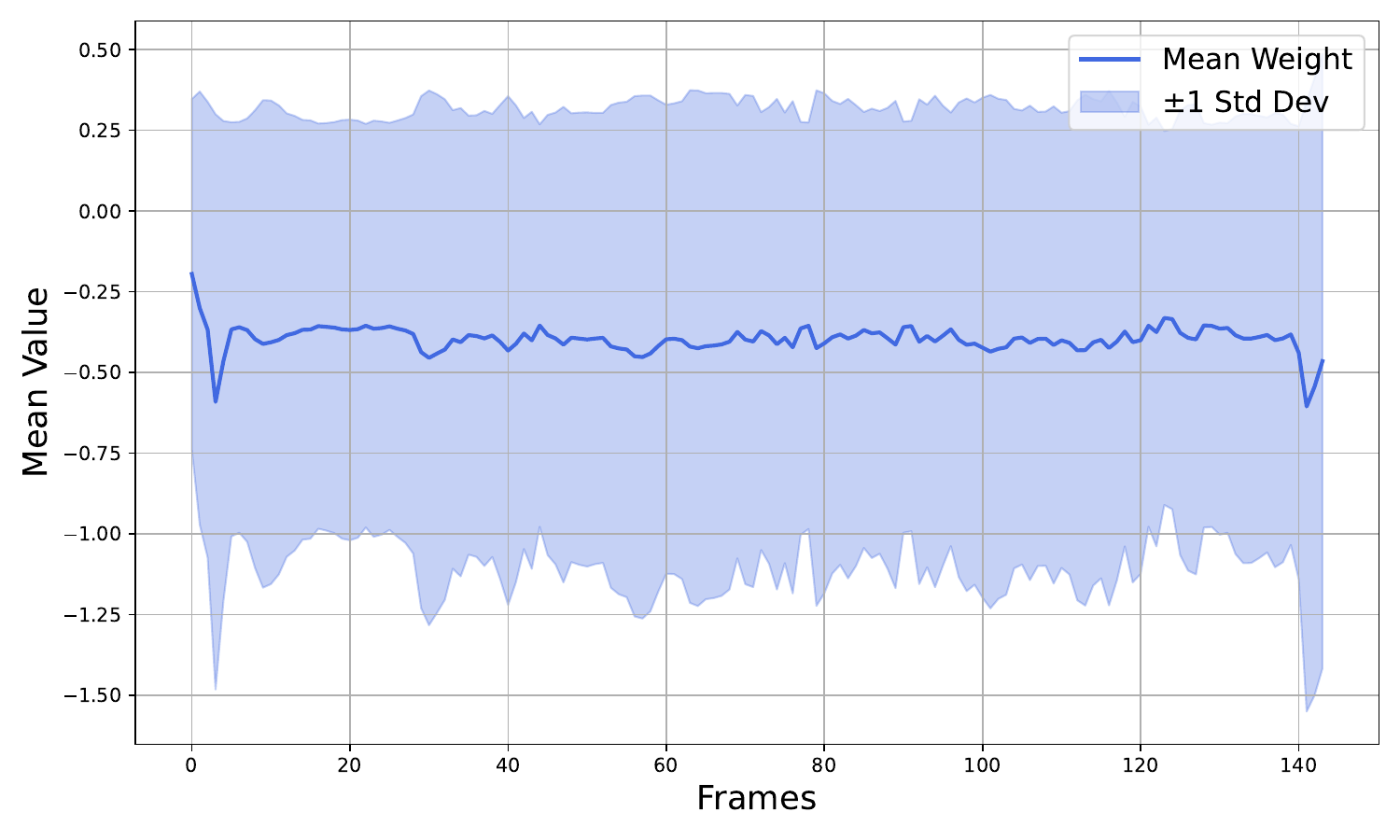}
  \caption{Frame-wise visualization of the weight variations generated by CAKG on the same sample as in Fig. \ref{fig:cakgv1}. The blue curve shows the average weight values, while the shaded region denotes ±1 standard deviation, illustrating the extent of fluctuation across frames.}
 \label{fig:cakgv2}
\end{figure}

\textbf{Gradient distribution comparison before and after using SR-CTC.} As shown in Fig. \ref{fig:gdistribution}, we compare the gradient distribution and zero-value ratio across feature extractor stages during the later training phase. The results demonstrate that SR-CTC effectively alleviates the sharp peak concentrated near zero and significantly reduces the proportion of zero gradients. Specifically, in stage 1, the proportion of zero gradients drops from 3.74\% to 0.64\% with SR-CTC. For stages 2, 3, and 4, the proportion falls below 0.1\%. Moreover, in stages 2 and 3, the gradient distribution becomes noticeably smoother. These findings confirm that SR-CTC mitigates the gradient vanishing problem in shallow layers during the later stages of training, facilitating parameter updates and helping the CTC loss escape suboptimal alignment paths.

\textbf{Visualization of Dynamic Convolution.} Fig.~\ref{fig:cakgv1} presents a visualization of a subset of dynamic convolution weights generated by CAKG during inference on a sample video. Each subplot corresponds to a specific frame, where the horizontal axis represents the temporal context dimension of the convolutional kernel, and the vertical axis indicates the mean weight value after channel-wise average pooling.  A noticeable observation is the stark contrast between the weight curves of boundary frames and those of middle frames. Specifically, Frame~0 and Frame~143 (the starting and ending frames) exhibit skewed or monotonic weight distributions, indicating that the model adapts to varying context lengths or information asymmetry—such as insufficient temporal context at the boundaries. This behavior demonstrates a clear instance of position-aware dynamic adaptation. While the middle-frame curves appear structurally similar, subtle variations exist in local values, peak positions, and minima, suggesting that the convolution kernels slightly adjust their attention within the receptive field depending on frame-specific context. This indicates a context-dependent modulation of the convolutional weights. Fig.~\ref{fig:cakgv2} illustrates the frame-wise variation of average kernel weights across the video. The variance is significantly higher at the beginning and end of the sequence (i.e., frames with index $<$10 or $>$135), as shown by the broader standard deviation band—especially in the negative direction. This is reasonable, as boundary frames have limited access to surrounding context, requiring the model to adjust more flexibly. For the rest of the sequence, the average weights remain relatively stable, with values fluctuating modestly between -0.25 and -0.4. These findings demonstrate that DCAC produces relatively stable weights across time, while maintaining the ability to adapt dynamically based on contextual information. Together, these visualizations highlight CAKG’s capability to generate frame-sensitive convolutional weights that adapt to local temporal context in a dynamic and position-aware manner.

\textbf{Qualitative comparison between  DESign and previous methods.}
As shown in Tab. \ref{tab:qual}, we present a comparison of recognition results between DESign and existing methods on four sample cases, with the WER of each result indicated at the end. It can be observed that DESign correctly identifies glosses that previous methods struggle with. For instance, in example (a), all prior methods fail to avoid the deletion error of ‘$\texttt{\_\_ON\_\_}$’, whereas DESign successfully recognizes this gloss. In example (b), previous methods produce substitution and deletion errors for ‘NOCHEINMAL’, while DESign correctly resolves them. These results highlight the robustness of DESign in handling challenging gloss recognition scenarios.

  

\begin{CJK*}{UTF8}{gbsn}
\begin{table*}[ht!]

\centering

\caption{Qualitative comparision between our DESign and previous methods (SEN \cite{sen}, TLP \cite{tlp}, CorrNet \cite{corrnet}, CorrNet+ \cite{corrnet+} ) on PHOENIX14. We use different colors to represent \sub{substitutions}, \del{deletions}, and \ins{insertions}, respectively.}
\label{tab:qual} 
\resizebox{\linewidth}{!}{
\begin{tabular}{l|l|c}
\toprule
\textbf{Example (a)} & & WER \\
\midrule
\multirow{2}{*}{Groundtruth} &\texttt{\_\_ON\_\_} IM-VERLAUF MITTE AUCH WOLKE AUFLOESEN WIND SCHWACH MAESSIG & \multirow{2}{*}{-} \\
& (On During Middle Also Cloud Dissolve Wind Weak Moderate) & \\
\midrule
\multirow{2}{*}{Pred. (SEN)} & \sub{******} IM-VERLAUF MITTE AUCH WOLKE \sub{VERSCHWINDEN} WIND SCHWACH MAESSIG & \multirow{2}{*}{22.2\%} \\
& (\sub{******} During Middle Also Cloud \sub{Vanish} Wind Weak Moderate) & \\
\midrule
\multirow{2}{*}{Pred. (TLP)} & \sub{******} IM-VERLAUF MITTE AUCH WOLKE \sub{VERSCHWINDEN} WIND SCHWACH MAESSIG & \multirow{2}{*}{22.2\%} \\
& (\sub{******} During Middle Also Cloud \sub{Vanish} Wind Weak Moderate) & \\
\midrule
\multirow{2}{*}{Pred. (CorrNet)} & \sub{******} IM-VERLAUF MITTE AUCH WOLKE AUFLOESEN WIND SCHWACH MAESSIG & \multirow{2}{*}{11.1\%} \\
& (\sub{******} During Middle Also Cloud Dissolve Wind Weak Moderate) & \\
\midrule
\multirow{2}{*}{Pred. (CorrNet$+$)} & \sub{******} IM-VERLAUF MITTE AUCH WOLKE \sub{VERSCHWINDEN} WIND SCHWACH MAESSIG & \multirow{2}{*}{22.2\%} \\
& (\sub{******} During Middle Also Cloud \sub{Vanish} Wind Weak Moderate) & \\
\midrule
\multirow{2}{*}{Pred. (DESign)} & \texttt{\_\_ON\_\_} IM-VERLAUF MITTE AUCH WOLKE AUFLOESEN WIND SCHWACH MAESSIG & \multirow{2}{*}{0.0\%} \\
& (\texttt{\_\_ON\_\_} During Middle Also Cloud Dissolve Wind Weak Moderate) & \\
\midrule

\midrule
\textbf{Example (b)} & & WER \\
\midrule
\multirow{2}{*}{Groundtruth} & \texttt{\_\_ON\_\_} WENN HIMMEL KLAR DOCH MEISTENS KALT FROST & \multirow{2}{*}{-} \\
& (\texttt{\_\_ON\_\_} When Sky Clear But Mostly COLD Frost) & \\
\midrule
\multirow{2}{*}{Pred. (SEN)} & \texttt{\_\_ON\_\_} WENN HIMMEL KLAR \del{****} \sub{NOCHEINMAL}  KALT FROST & \multirow{2}{*}{25.0\%} \\
& (On When Sky Clear \del{****} \sub{Night} Cold Frost) & \\
\midrule
\multirow{2}{*}{Pred. (TLP)} &  \texttt{\_\_ON\_\_} WENN HIMMEL KLAR \del{****} \del{*******}  KALT FROST & \multirow{2}{*}{25.0\%} \\
& (On When Sky Clear \del{****} \del{*******} Cold Frost) & \\
\midrule
\multirow{2}{*}{Pred. (CorrNet)} & \texttt{\_\_ON\_\_} WENN HIMMEL KLAR \del{****} \del{********} \sub{SOLL} FROST & \multirow{2}{*}{37.5\%} \\
& (\texttt{\_\_ON\_\_} When Sky Clear \del{****} \del{*******} \sub{Should} Frost) & \\
\midrule
\multirow{2}{*}{Pred. (CorrNet+)} & \texttt{\_\_ON\_\_} WENN HIMMEL KLAR  DOCH \del{********} \sub{KOENNEN} FROST & \multirow{2}{*}{25.0\%} \\
& (\texttt{\_\_ON\_\_} When Sky Clear But \del{********} \sub{Can} Frost) & \\
\midrule
\multirow{2}{*}{Pred. (DESign)} & \texttt{\_\_ON\_\_} WENN HIMMEL KLAR DOCH MEISTENS \del{****} FROST & \multirow{2}{*}{12.5\%} \\
& (\texttt{\_\_ON\_\_} When Sky Clear But Mostly \del{****} Frost) & \\

\midrule

\midrule

\textbf{Example (c)} & & WER \\
\midrule
\multirow{2}{*}{Groundtruth} & FRISCH MORGEN ANGENEHM DEUTSCH LAND ABER IM-MOMENT DANN VIEL REGEN REGION OST REGION OST LANGSAM & \multirow{2}{*}{-} \\
& (Fresh Morning Pleasant German Land But In-Moment Then Much Rain Region East Region East Slowly) & \\
\midrule
\multirow{2}{*}{Pred. (SEN)} & FRISCH MORGEN \sub{MAXIMAL} DEUTSCH LAND \sub{AUCH} IM-MOMENT DANN VIEL REGEN REGION OST \del{******} \del{***} \sub{KOMMEN} \ins{VERSCHWINDEN}& \multirow{2}{*}{37.5\%} \\
& (Fresh Morning \sub{Maximal} German Land But In-Moment Then Much Rain Region East \del{******} \del{***} \sub{Come} \ins{Disappear}) & \\
\midrule
\multirow{2}{*}{Pred. (TLP)} & FRISCH MORGEN \sub{WEITER} DEUTSCH LAND ABER \sub{ENGLAND} DANN VIEL REGEN \sub{KOMMEN} OST \sub{KOMMEN} OST LANGSAM \ins{VERSCHWINDEN}& \multirow{2}{*}{31.3\%} \\
& (Fresh Morning \sub{Continue} German Land But \sub{England} Then Much Rain \sub{Come} East \sub{Come} East Slowly \ins{Disappear}) & \\
\midrule
\multirow{2}{*}{Pred. (CorrNet)} & FRISCH MORGEN ANGENEHM \sub{TEMPERATUR} LAND ABER IM-MOMENT \del{****} VIEL REGEN \sub{KOMMEN} OST \del{******} OST LANGSAM \ins{VERSCHWINDEN}& \multirow{2}{*}{29.4\%} \\
& (Fresh Morning Pleasant \sub{Temprature} Land But In-Moment \del{****} Much Rain \sub{Come} East \del{******} East Slowly \ins{Disappear}) & \\
\midrule
\multirow{2}{*}{Pred. (CorrNet+)} & FRISCH MORGEN ANGENEHM DEUTSCH LAND ABER \del{*********} DANN VIEL REGEN \sub{KOMMEN} OST \sub{KOMMEN} OST LANGSAM \ins{VERSCHWINDEN}& \multirow{2}{*}{25.0\%} \\
& (Fresh Morning Pleasant German Land But \del{*********} Then Much Rain \sub{Come} East \sub{Come} East Slowly \ins{Disappear}) & \\
\midrule
\multirow{2}{*}{Pred. (DESign)} & FRISCH MORGEN ANGENEHM DEUTSCH LAND ABER IM-MOMENT DANN VIEL REGEN \sub{BEWOELKT} OST \sub{KOMMEN} OST LANGSAM \ins{VERSCHWINDEN} & \multirow{2}{*}{18.8\%} \\
& (Fresh Morning Pleasant German Land But In-Moment Then Much Rain \sub{Cloudy} East \sub{Come} East Slowly \ins{Disappear}) & \\

\midrule
\textbf{Example (d)} & & WER \\
\midrule
\multirow{2}{*}{Groundtruth} &SUED VIERHUNDERT SIEBENHUNDERT METER HOHE MÖGLICH SCHNEE ENORM& \multirow{2}{*}{-} \\
& (South Four-Hundred Seven-Hundred Meter Height Possible Snow Enormous) & \\
\midrule
\multirow{2}{*}{Pred. (SEN)} &SUED \ins{REGION} \sub{SECHSHUNDERT} SIEBENHUNDERT METER HOHE MÖGLICH SCHNEE \sub{MOEGLICH}& \multirow{2}{*}{37.5\%} \\
& (South \ins{Region} \sub{Six-Hundred} Seven-Hundred Meter Height Possible Snow \sub{Possible}) & \\
\midrule
\multirow{2}{*}{Pred. (TLP)} &  SUED \ins{REGION} VIERHUNDERT \sub{ACHTHUNDERT} METER HOHE MÖGLICH SCHNEE \sub{MOEGLICH}& \multirow{2}{*}{33.3\%} \\
& (South \ins{Region} Four-Hundred \sub{Eight-Hundred} Meter Height Possible Snow Enormous \sub{Possible}) & \\
\midrule
\multirow{2}{*}{Pred. (CorrNet)} & SUED \ins{REGION} \sub{FRANKREICH} SIEBENHUNDERT METER HOHE MÖGLICH SCHNEE ENORM \sub{MOEGLICH}& \multirow{2}{*}{33.3\%} \\
& (South \ins{Region} \sub{France} Seven-Hundred Meter Height Possible Snow Enormous \sub{Possible}) & \\
\midrule
\multirow{2}{*}{Pred. (CorrNet+)} & SUED \ins{REGION} VIERHUNDERT \sub{ACHTHUNDERT} METER HOHE MÖGLICH SCHNEE ENORM & \multirow{2}{*}{22.2\%} \\
& (South \ins{Region} Four-Hundred 
\sub{Eight-Hundred} Meter Height Possible Snow Enormous) & \\
\midrule
\multirow{2}{*}{Pred. (DESign)} & SUED \ins{REGION} VIERHUNDERT SIEBENHUNDERT METER HOHE MÖGLICH SCHNEE ENORM & \multirow{2}{*}{11.1\%} \\
& (South \ins{Region} Four-Hundred Seven-Hundred Meter Height Possible Snow Enormous) & \\

\bottomrule
\end{tabular}}

\end{table*}
\end{CJK*}

\section{Limitations}
\label{sec:7}
Despite the effectiveness of our proposed methods, several limitations remain. First, although DCAC's frame-wise kernel generation greatly enhances modeling capability, we observe that adjacent frames often exhibit minimal differences—especially during slow-motion sequences. Introducing a selective kernel-sharing strategy could potentially yield comparable performance while reducing computational overhead. Additionally, SR-CTC, like previous methods \cite{vac, smkd, radialctc, c2st}, does not fundamentally resolve the inherent peaking issue of CTC. Rather, it acts as an efficient mitigation approach. While it alleviates the severity of peaking to a certain extent, the problem may still persist. These insights point to exciting avenues for future research. Enhancing DCAC with adaptive kernel-sharing mechanisms and developing more principled solutions to address the CTC peaking phenomenon could further improve model efficiency and robustness.

\end{document}